%% file: main.tex
\documentclass[11pt]{article}

\usepackage{fullpage}
\usepackage{times}
\usepackage{url}
\usepackage{latexsym}
\usepackage{amsfonts}
\usepackage{multirow}
\usepackage{hyperref}
\hypersetup{
  colorlinks = true,
  citecolor = blue
}
\usepackage{booktabs}
\usepackage{mathtools}
\usepackage{enumitem}
\usepackage{color}
\usepackage{algorithm,algorithmic, amsmath, amssymb, amsthm}
\usepackage{adjustbox}

\newtheorem{theorem}{Theorem}[section]
\newtheorem{lemma}[theorem]{Lemma}

\allowdisplaybreaks

\newenvironment{definition}[1][Definition]{\begin{trivlist}
\item[\hskip \labelsep {\bfseries #1}]}{\end{trivlist}}

\newenvironment{remark}[1][Remark]{\begin{trivlist}
\item[\hskip \labelsep {\bfseries #1}]}{\end{trivlist}}

\newcommand{\inner}[1]{\left\langle#1\right\rangle}

\def\R{\mathbb{R}}

\def\bone{{\mathbf 1}}

\def \bp {{\mathbf{p}}}
\def \bq {{\mathbf{q}}}

\def\bw{{\mathbf w}}
\def\bx{{\mathbf x}}
\def\by{{\mathbf y}}
\def\bz{{\mathbf z}}

\def\bC{\mathbf C}

\def\bK{\mathbf K}

\def\calX{\mathcal X}
\def\calY{\mathcal Y}

\def\bS{\mathbf S}

\def\argmax{\mathop{\rm arg\,max}\limits}

\def\maxop{\mathop{\rm max}\limits} 
\def\minop{\mathop{\rm min}\limits}

\def \bwP{{\bw_{P}}}

\newcommand{\eat}[1]{}

\newcommand\norm[1]{\left\lVert#1\right\rVert}

\newcommand*\samethanks[1][\value{footnote}]{\footnotemark[#1]}

\newcommand{\spots}{SPOTsimple}
\newcommand{\spotg}{SPOTgreedy}

\begin{document}

\title{SPOT: A framework for selection of prototypes\\ using optimal transport}

\author{Karthik S. Gurumoorthy\thanks{Equal contribution.}\ \ \thanks{India Machine Learning, Amazon, India. Email: gurumoor@amazon.com.} \quad Pratik Jawanpuria\samethanks[1]\ \ \thanks{Microsoft, India. Email: \{pratik.jawanpuria,bamdevm\}@microsoft.com.} \quad Bamdev Mishra\samethanks[1]\ \ \samethanks[3]}

\date{}

\maketitle

\input{abstract}


\input{manuscript}

\input{supplementary}

\bibliography{references}
\bibliographystyle{amsalpha}

\end{document}

%% file: abstract.tex
\begin{abstract}
In this work, we develop an optimal transport (OT) based framework to select informative prototypical examples that best represent a given target dataset. Summarizing a given target dataset via representative examples is an important problem in several machine learning applications where human understanding of the learning models and underlying data distribution is essential for decision making. We model the prototype selection problem as learning a sparse (empirical) probability distribution having the minimum OT distance from the target distribution. The learned probability measure supported on the chosen prototypes directly corresponds to their importance in representing the target data. We show that our objective function  enjoys a key property of submodularity and propose an efficient greedy method that is both computationally fast and possess deterministic approximation guarantees. Empirical results on several real world benchmarks illustrate the efficacy of our approach. 



\end{abstract}

%% file: manuscript.tex
\section{Introduction}
Extracting informative and influential samples that best represent the underlying data-distribution is a fundamental problem in machine learning \cite{weiser1982programmers,sproto,kim14a,koh17,yeh18a}. As sizes of datasets have grown, summarizing a dataset with a collection of representative samples from it is of increasing importance to data scientists and domain-specialists \cite{bien11a}. 
Prototypical samples offer interpretative value in every sphere of humans decision making where machine learning models have become integral such as healthcare \cite{caruana}, information technology \cite{irt}, and entertainment \cite{lime}, to name a few. 
In addition, extracting such compact synopses play a pivotal tool in depicting the scope of a dataset, in detecting outliers \cite{Kim16}, and for compressing and manipulating data distributions \cite{rousseeuw09a}. 
Going across domains to identify representative examples from a source set that explains a different target set have recently been applied in model agnostic PU learning \cite{PULearning}.
Existing works \cite{lozano06,wei15a} have also studied the generalization properties of machine learning models trained on a prototypical subset of a large dataset. 

Works such as \cite{sproto,crammer02a,wohlhart13a,wei15a} consider selecting representative elements (henceforth also referred to as \textit{prototypes}) in the supervised setting, i.e., the selection algorithm has access to the label information of the data points. Recently \cite{Kim16,proto} have also explored the problem of prototype selection in the unsupervised setting, in which the selection algorithm has access only to the feature representation of the data points. They view the given dataset $Y$ and a candidate prototype set $P$ (subset of a source dataset $X$) as empirical distributions $q$ and $p$, respectively. The prototype selection problem, therefore, is modeled as searching for a distribution $p$ (corresponding to a set $P \subset X$ of data points, typically with a small cardinality) that is a good approximation of the distribution $q$. For example, \cite{Kim16,proto} employ the maximum mean discrepancy (MMD) distance \cite{Gretton06} to measure the similarity between the two distributions. 

It is well-known that the MMD induces the ``flat'' geometry of reproducing kernel Hilbert space (RKHS) on the the space of probability distributions as it measures the distance between the mean embeddings of distributions in the RKHS of a universal kernel \cite{smola07a,Gretton06,gretton12a}). The individuality of data points is also lost while computing distance between mean embeddings in the MMD setting. The optimal transport (OT) framework, on the other hand, provides a natural metric for comparing probability distributions while respecting the underlying geometry of the data \cite{villani09a,peyre19a}.
Over the last few years, OT distances (also known as the Wasserstein distances) have found widespread use in several machine learning applications such as image retrieval \cite{rubner00}, shape interpolation \cite{solomon15a}, domain adaptation \cite{courty17a}, supervised learning \cite{frogner15a}, and generative model training \cite{arjovsky17a}, among others. 
The \textit{transport plan}, learned while computing the OT distance between the source and target distributions, is the joint distribution between the source and the target distributions. Compared to the MMD, the OT distances enjoy several advantages such as being faithful to the ground metric (geometry over the space of probability distributions) and identifying correspondences at the fine grained level of individual data points via the transport plan. 

In this paper, we focus on the unsupervised prototype selection problem and view it from the perspective of the optimal transport theory. To this end, we propose a novel framework for {\bf S}election of {\bf P}rototypes using the {\bf O}ptimal {\bf T}ransport theory or the {\bf SPOT} framework for searching a subset $P$ from a source dataset $X$ (i.e., $P \subset X$) that best represents a target set $Y$. 
We employ the Wasserstein distance to estimate the closeness between the distribution representing a candidate set $P$ and set $Y$. Unlike the typical OT setting, the source distribution (representing $P$) is unknown in SPOT and needs to be learned along with the transport plan. The prototype selection problem is modeled as learning an empirical source distribution $p$ (representing set $X$) that has the minimal Wasserstein distance with the empirical target distribution (representing set $Y$). Additionally, we constrain $p$ to have a small support set (which represents $P\subset X$). The learned distribution $p$ is also indicative of the relative importance of the prototypes in $P$ in representing $Y$. Our main contributions are as follows.
\begin{itemize}
\item We propose a novel prototype selection framework, SPOT, based on the OT theory. 
\item We prove that the objective function of the proposed optimization problem in SPOT is submodular, which leads to a tight approximation guarantee of $\left(1-e^{-1}\right)$ using greedy approximation algorithms~\cite{Nemhauser78}. The computations in the proposed greedy algorithm can be implemented efficiently.
\item We explain the popular k-medoids clustering~\cite{rousseeuw09a} formulation as a special case of SPOT formulation (when the source and the target datasets are the same). We are not aware of any prior work that describes such a connection though the relation between Wasserstein distance minimization and k-means is known~\cite{kmeansOT,cuturi14a}.
\item Our empirical results show that the proposed algorithm outperforms existing baselines on several real-world datasets. The optimal transport framework allows our approach to seamlessly work in settings where the source ($X$) and the target ($Y$) datasets are from different domains. 
\end{itemize}

The outline of the paper is as follows. We provide a brief review of the optimal transport setting, the prototype selection setting, and key definitions in the submodular optimization literature in Section~\ref{sec:background}. 
The proposed SPOT framework and algorithms are presented in Section~\ref{sec:proposed}. We discuss how SPOT relates to existing works in Section~\ref{sec:related}. The empirical results are presented in Section~\ref{sec:experiment}. We conclude the paper in Section~\ref{sec:conclusion}. The proofs and additional results on datasets are presented in the appendix. 

\section{Background}\label{sec:background}
\eat{We begin by discussing the optimal transport problem.}

\subsection{Optimal transport (OT)}
Let $X\coloneqq\{\bx_i\}_{i=1}^m$ and $Y\coloneqq\{\by_j\}_{j=1}^n$ be i.i.d. samples from the source and the target distributions $p$ and $q$, respectively. In several applications, the true  distributions are generally unknown. However, their empirical estimates exist and can be employed as follows:
\begin{equation}\label{eqn:empirical-distribution}
p\coloneqq \sum_{i=1}^m \bp_i\delta_{\bx_i},\;\;\;\;\; q\coloneqq \sum_{j=1}^n \bq_j\delta_{\by_j},
\end{equation}
where the probability associated with samples $\bx_i$ and $\by_j$ are $\bp_i$ and $\bq_j$, respectively, and $\delta$ is the Dirac delta function. The vectors $\bp$ and $\bq$ lie on simplices $\Delta_{m}$ and $\Delta_n$, respectively, where $\Delta_{k}\coloneqq\{\bz\in\R_{+}^k|\sum_i \bz_i=1\}$. The OT problem \cite{kantorovich42a} aims at finding a transport plan $\gamma$ (with the minimal transporting effort) as a solution to
\begin{equation}\label{eqn:emd}
\minop_{\gamma\in\Gamma(\bp,\bq)} \inner{\bC,\gamma},
\end{equation}
where $\Gamma(\bp,\bq)\coloneqq\{\gamma\in\R_{+}^{m\times n}|\gamma\bone=\bp;\gamma^\top\bone=\bq\}$ is the space of joint distribution between the source and the target marginals. Here, $\bC\in\R_{+}^{m\times n}$ is the ground metric computed as $\bC_{ij}=c(\bx_i,\by_j)$ and the function $c:\calX \times \calY \rightarrow \R_{+}:(\bx,\by)\rightarrow c(\bx,\by)$ represents the \textit{cost} of transporting a unit mass from source $\bx\in\calX$ to target $\by\in\calY$.

The optimization problem (\ref{eqn:emd}) is a linear program. Recently, \cite{cuturi13a} proposed an efficient solution for learning entropy regularized transport plan $\gamma$ in (\ref{eqn:emd}) using the Sinkhorn algorithm \cite{knight08a}. For a recent survey on OT with focus on machine learning applications, please refer to \cite{peyre19a}. 

\subsection{Prototype selection}
Selecting representative elements is often posed as identifying a subset $P$ of size $k$ from a set of items $X$ (e.g., data points, features, etc.). The quality of selection is usually governed via a scoring function $f(P)$, which encodes the desirable properties of prototypical samples. For instance, in order to obtain a compact yet informative subset $P$, the scoring function should discourage redundancy. Recent works \cite{Kim16,proto} have posed prototype selection within the submodular optimization setting by maximizing a MMD based scoring function on the weights ($\bw$) of the prototype elements:
\begin{equation}
\label{eqn:MMDObjective}
l(\bw) = \boldsymbol{\mu}^T \bw - \frac{1}{2} \bw^T \bK \bw \mbox{ s.t. }  \|\bw\|_0 \leq k.
\end{equation}
Here, $\|\bw\|_0$ is $\ell_0$ norm of $\bw$ representing the number of non-zero values, 
the entries of the vector $\boldsymbol{\mu}$ contains the mean of the inner product for every source point with the target data points computed in the kernel embedding space, 
and $\bK$ is the Gram matrix of a universal kernel (e.g., Gaussian) corresponding to the source instances. 
The locations of non-zero values in $\bw$, $supp(\bw) = \{i: \bw_i > 0\}$, known as its \emph{support} correspond to the element indices that are chosen as prototypes, i.e. $P = supp(\bw)$. While the MMD-Critic method in \cite{Kim16} enforces that all non-zero entries in $\bw$ equal to $1/k$, the ProtoDash algorithm in \cite{proto} imposes non-negativity constraints and learns $\bw$ as part of the algorithm. Both propose greedy algorithms that effectively evaluate the \textit{incremental} benefit of adding an element in the prototypical set $P$. In contrast to the MMD function in (\ref{eqn:MMDObjective}), to the best of our knowledge, ours is the first work which leverages the optimal transport (OT) framework to extract such compact representation. We prove that the proposed objective function is submodular, which ensures tight approximation guarantee using greedy approximate algorithms. 

\subsection{Submodularity}
\label{sec:submodularity}
We briefly review the concept of submodular and weakly submodular functions, which we later use to prove key theoretical results. 
\begin{definition}[Submodularity and Monotonicity]
\label{def:sub}
Consider any two sets $A \subseteq B \subseteq [m]$. A set function $f(.)$ is \emph{submodular} if and only if for any $i \notin B$, $f\left(A \cup {i}\right) - f(A) \geq f\left(B \cup {i}\right) - f(B)$. The function is called \emph{monotone} when $f(A)\leq f(B)$.
\end{definition}
Submodularity implies diminishing returns where the incremental gain in adding a new element $i$ to a set $A$ is at least as high as adding to its superset $B$~\cite{fujishige05}. Another characterization of submodularity is via the submodularity ratio \cite{weaksub,weaksubInit} defined as follows.
\begin{definition}[Submodularity Ratio]
Given two disjoint sets $L$ and $S$, and a set function $f(\cdot)$, the submodularity ratio of $f(\cdot)$ for the ordered pair ($L,S$) is given by:
\begin{equation}\label{eqn:subratio}
\alpha_{L,S} \coloneqq \frac{\sum\limits_{i \in S} \left[f\left(L\cup \{i\}\right) - f(L)\right]}{f\left(L \cup S\right)-f(L)}.
\end{equation}
\end{definition}
Submodularity ratio captures the increment in $f(\cdot)$ by adding the entire subset $S$ to $L$, compared to summed gain of adding its elements individually to $L$. It is known that $f(\cdot)$ is submodular if and only if $\alpha_{L,S} \geq 1, \forall L,S$. In the case where $0\leq \epsilon\leq\alpha_{L,S} <1$ for an independent constant $\epsilon$, $f(\cdot)$ is called \textit{weakly submodular}~\cite{weaksubInit}. 

We define submodularity ratio of a set $P$ with respect to an integer $s$ as follows:
\begin{equation}
\label{eqn:subratioPs}
\alpha_{P,s} \coloneqq \max_{{L,S: L \cap S= \emptyset, L \subseteq P, |S|\leq s}} \alpha_{L,S}.
\end{equation}
It should be emphasized that unlike the definition in~\cite[Equation 3]{weaksub}, the above Equation (\ref{eqn:subratioPs}) involves the \texttt{max} operator instead of the \texttt{min}. This specific form is later used to produce approximation bounds for the proposed approach (presented in Algorithm~\ref{alg:SPOT-Greedy}). Both (strongly) submodular and weakly submodular functions enjoy provable performance bounds when the set elements are selected incrementally and greedily~\cite{Nemhauser78,weaksub,proto}.
\eat{
\begin{remark}
\label{remark:alphaPs}
Let $|S|=s$. When $f(.)$ is monotone, then 
\begin{align}
\label{bound:alphaLS}
\alpha_{L,S} \leq \frac{\sum\limits_{i \in S} \left[f\left(L\cup \{i\}\right) - f(L)\right]}{\max\limits_{i \in S} \left[f\left(L\cup \{i\}\right) - f(L)\right]} \leq s
\end{align}
and hence $\alpha_{P,s} \leq s$. Particularly when $s=1$, $\alpha_{P,1} = 1$ as for any $L \subseteq P$, $\alpha_{L,S}=1$ when $|S|=1$. 
\end{remark}
}

\section{SPOT framework}\label{sec:proposed}


\eat{We begin by discussing how to pose prototype selection as an OT problem.}

\subsection{SPOT problem formulation}\label{subsec:problem}
Let $X = \{\bx_i\}_{i=1}^m$ be a set of $m$ source points, $Y = \{\by_j\}_{j=1}^n$ be a target set of $n$ data points, and $\bC\in\R_{+}^{m\times n}$ represents the ground metric. Our aim is to select a small and weighted subset $P \subset X$ of size $k \ll m$ that best describes $Y$. To this end, we develop an optimal transport (OT) based framework for selection of prototypes.

Traditionally, OT is defined as a minimization problem over the transport plan $\gamma$ as in (\ref{eqn:emd}). In our setting, we pre-compute a \textit{similarity} matrix $\bS\in\R_{+}^{m\times n}$ from $\bC$, for instance, as $\bS_{ij} = \beta - \bC_{ij}$ where $\beta > \norm{\bC}_{\infty}$. This allows to equivalently represent the OT problem (\ref{eqn:emd}) as a maximization problem with the objective function as $\inner{\bS,\gamma}$. 
Treating it as a maximization problem enables to establish connection with submodularity and leverage standard greedy algorithms for its optimization \cite{Nemhauser78}. 

We pose the problem of selecting {a prototypical set (of utmost size $k$)} as learning a sparse support empirical source distribution {$w = \sum_{\bx_i \in P} \bw_i \delta_{\bx_i}$} that has {maximum} closeness to the target distribution in terms of the optimal transport measure. Here, the weight $\bw\in\Delta_m$, where $\Delta_{m}\coloneqq\{\bz\in\R_{+}^m|\sum_i \bz_i=1\}$. Consequently, $\bw$ denotes the relative importance of the samples. Hence, the constraint $|P|\leq k$ for the prototype set $P$ translates to $|supp(\bw)|\leq k$ {where $supp(\bw) \subseteq P$}. 

We evaluate the suitability of a candidate prototype set $P\subset X$ with an OT based measure on sets. To elaborate, index the elements in $X$ from $1$ to $m$ and let $[m]\coloneqq \{1,2,\ldots,m\}$ denote the first $m$ natural numbers. Given any index set of prototypes $P \subseteq [m]$, define a set function $f: 2^{[m]} \rightarrow \R_{+}$ as:
\begin{equation}
\label{eqn:f}
f(P) \coloneqq \maxop_{\bw: supp(\bw) \subseteq P}\  \maxop_{\gamma\in\Gamma(\bw,\bq)} \inner{\bS,\gamma},
\end{equation}
where $\bq\in\Delta_n$ corresponds to the (given) weights of the target samples\footnote{In the absence of domain knowledge, uniform weights $\bq=\bone/n$ can be a default choice.} in the empirical target distribution $q$ as in (\ref{eqn:empirical-distribution}). The learned transport plan $\gamma$ in (\ref{eqn:f}) is a joint distribution between the elements in $P$ and $Y$, which may be useful in downstream applications requiring, e.g., barycentric mapping. 

Our goal is to find that set $P$ which maximizes $f(\cdot)$ subject to the cardinality constraint. To this end, the proposed SPOT problem is
\begin{equation}\label{eqn:prototype-selection}
   P^{*} = \argmax_{P\subseteq [m],|P|\leq k} f(P), 
\end{equation}
where $f(P)$ is defined in (\ref{eqn:f}). The entries of the optimal weight vector $\bw^{\ast}$ corresponding to $P^{*}$ in (\ref{eqn:prototype-selection}) indicate the importance of the prototypes in summarizing set $Y$. The SPOT (\ref{eqn:prototype-selection}) and the standard OT (\ref{eqn:emd}) settings are different as:
(a) the source distribution {$w$} is learned as a part of the SPOT optimization problem formulation and (b) the source distribution {$w$} is enforced to have a sparse support of utmost size $k$ so that the prototypes create a compact summary.

In the next section, we analyze the objective function in the SPOT optimization problem (\ref{eqn:prototype-selection}), characterize it with a few desirable properties, and develop a computationally efficient greedy approximation algorithm.
\subsection{Equivalent reduced representations of SPOT objective}
\label{subsec:optimization}
Though the definition of the scoring function $f(\cdot)$ in (\ref{eqn:f}) involves maximization over two coupled variables $\bw$ and $\gamma$, it can be reduced to an equivalent optimization problem involving only $\gamma$ (by eliminating $\bw$ altogether). To this end, let $k=|P|$ and denote $\bS_{P}$ a $k \times n$ sub-matrix of $\bS$ containing only those rows indexed by $P$. We then have the following lemma:
\begin{lemma}
\label{lemma:welimination}
The set function $f(\cdot)$ in (\ref{eqn:f}) can be equivalently defined as an optimization problem only over the transport plan, i.e.,
\begin{equation}
\label{eqn:fred}
f(P) = \maxop_{\gamma\in\Gamma_P(\bq)} \inner{\bS_{P},\gamma},    
\end{equation}
where $\Gamma_P(\bq)\coloneqq\{\gamma\in\R_{+}^{k\times n}|\gamma^\top\bone=\bq\}$. Let $\gamma^{*}$ be an optimal solution of (\ref{eqn:fred}). Then, $(\bw^{*},\gamma^{*})$ is an optimal solution of (\ref{eqn:f}) where $\bw^{*}=\gamma^{*}\bone$. 
\end{lemma}
A closer look into the set function in (\ref{eqn:fred}) reveals that the optimization for $\gamma$ can be done in parallel over the $n$ target points, and its solution assumes a closed-form expression. It is worth noting that the constraint $\gamma^T \bone = \bq$ as well as the objective $\inner{\bS_{P},\gamma}$ decouple over each column of $\gamma$. Hence, (\ref{eqn:fred}) can be solved across the columns of variable $\gamma$ independently, thereby allowing parallelism over the target set. In other words,
\begin{equation}
\label{eqn:fdecoupled}
f(P) = \sum\limits_{j=1}^n \maxop_{\gamma^j \in \mathbb{R}_+^k} \inner{\bS_{P}^j,\gamma^j}, \textup{  s.t.  } \bone^T\gamma^j = \bq_j\ \forall j,
\end{equation}
where $\bS_P^j$ and $\gamma^j$ denote the $j^{th}$ column vectors of the matrices $\bS_P$ and $\gamma$, respectively. Furthermore, if $i_j$ denotes the location of the maximum value in the vector $\bS_P^j$, then an optimal solution $\gamma^*$ can be easily seen to inherit an extremely sparse structure with exactly one non-zero element in each column $j$ at the row location $i_j$, i.e., $\gamma^*_{i_j, j}=\bq_j, \forall j$ and $0$ everywhere. So (\ref{eqn:fdecoupled}) can be reduced to
\begin{equation}
\label{eqn:reducedform}
f(P) = \sum\limits_{j=1}^n \bq_j \maxop_{i \in P}\bS_{ij}.
\end{equation}
The above observation makes the computation $f(P)$ in (\ref{eqn:reducedform}) particularly suited when using GPUs. In addition, due to this specific solution structure in (\ref{eqn:reducedform}), determining the function value for any incremental set is a relatively inexpensive operation as presented in our next result. 
\begin{lemma}[Fast incremental computation]
\label{lemma:incrementalf}
Given any set $P$ and its function value $f(P)$, the value at the incremental selection $f\left(P \cup S\right)$ obtained by adding $s = |S|$ new elements to $P$, can be computed in {$O(sn)$}.  \eat{If there are $n$ nodes working in parallel, $f\left(P \cup S\right)$ can be obtained in {$O(s)$} per node.}
\end{lemma}
\begin{remark}
By setting $P=\emptyset$ and $f(\emptyset)=0$, $f(S)$ for any set $S$ can be determined efficiently as discussed in Lemma~\ref{lemma:incrementalf}.
\end{remark}

\subsection{SPOT optimization algorithms}
As obtaining the global optimum subset $P^{\ast}$ for the problem (\ref{eqn:prototype-selection}) is NP complete, we now present two approximation algorithms for SPOT: {\spots} and {\spotg}.

\subsubsection{{\spots}: a fast heuristic algorithm.}
\label{sec:SPOT-Simple}
{\spots} is an extremely fast heuristic that works as follows. For every source point $\bx_i$, {\spots} determines the indices of target points $\mathcal{T}_i = \{ j : \bS_{ij} \geq \bS_{\tilde{i}j} \text{for all} \tilde{i} \not= i \}$ that have the highest similarity to $\bx_i$ compared to other source points. In other words, it solves (\ref{eqn:reducedform}) with $P=[m]$, i.e., no cardinality constraint, to determine the initial transport plan $\gamma$ where $\gamma_{ij} = \bq_j$ if $j\in \mathcal{T}_i$ and $0$ everywhere else. It then computes the source weights as $\bw = \gamma \bone$ with each entry $\bw_{i} = \sum\limits_{j \in \mathcal{T}_i} \bq_j$. The top-$k$ source points based on the weights $\bw$ are chosen as the prototype set $P$. The final transport plan $\gamma_P$ is recomputed using (\ref{eqn:reducedform}) over $P$. The total computational cost incurred by {\spots} for selecting $k$ prototypes is $O(mn)$.

\subsubsection{{\spotg}: a greedy and incremental prototype selection algorithm.}
As we discuss later in our experiments (section~\ref{sec:experiment}), though {\spots} is computationally very efficient, its accuracy of prototype selection is sensitive to the skewness of class instances in the target distribution. When the samples from different classes are uniformly represented in the target set, {\spots} is indeed able to select prototypes from the source set that are representative of the target. However, when the target is skewed and the class distributions are no longer uniform, {\spots} primarily chooses from the dominant class leading to biased selection and poor performance (see Figure \ref{fig:accuracy_vs_prop}(a)). 

To this end, we present our method of choice {\spotg}, detailed in Algorithm~\ref{alg:SPOT-Greedy}, that leverages the following desirable properties of the function $f(\cdot)$ in (\ref{eqn:reducedform}) to \emph{greedily and incrementally} build the prototype set $P$. For choosing $k$ protototypes, {\spotg} costs $O(mnk/s)$. As most operations in {\spotg} involve basic matrix manipulations, the practical implementation cost of {\spotg} is considerably low.

\begin{lemma}[Submodularity]
\label{thm:submodular}
The set function $f(\cdot)$ defined in (\ref{eqn:reducedform}) is monotone and submodular \cite{mirzasoleiman16a}.
\end{lemma}
The submodularity of $f(.)$ enables to provide provable approximation bounds for greedy element selections in {\spotg}. The algorithm begins by setting the current selection $P=\emptyset$. Without loss of generality, we assume $f(\emptyset)=0$ as $f(\cdot)$ is monotonic. In each iteration, it determines those $s$ elements from the remainder set $[m]\setminus P$, denoted by {$S$}, that when individually added to $P$ result in maximum incremental gain. {This can be implemented efficiently as discussed in Lemma~\ref{lemma:incrementalf}.} 
Here $s \geq 1$ is the user parameter that decides the number of elements chosen in each iteration. The set $S$ is then added to $P$. The algorithm proceeds for $\lceil \frac{k}{s} \rceil$ iterations to select $k$ prototypes. As function $f(\cdot)$ in (\ref{eqn:fred}) is both monotone and submodular, it has the characteristic of diminishing returns. Hence, an alternative stopping criterion could be the minimum expected increment $\epsilon$ in the function value at each iteration. The algorithm stops when the increment in the function value is below the specified threshold $\epsilon$.
 
\textbf{Approximation guarantee for {\spotg}}. We note the following result on the upper bound on the submodularity ratio (\ref{eqn:subratio}). 
Let $s=|S|$. When $f(\cdot)$ is monotone, then 
\begin{align}
\label{bound:alphaLS}
\alpha_{L,S} \leq \frac{\sum\limits_{i \in S} \left[f\left(L\cup \{i\}\right) - f(L)\right]}{\max\limits_{i \in S} \left[f\left(L\cup \{i\}\right) - f(L)\right]} \leq s
\end{align}
and hence $\alpha_{P,s} \leq s$. 
In particular, $s=1$ implies $\alpha_{P,1} = 1$, as for any $L \subseteq P$, $\alpha_{L,S}=1$ when $|S|=1$. 
Our next result provides the performance bound for the proposed {\spotg} algorithm. 
\begin{theorem}[Performance bounds for {\spotg}]
\label{thm:bounds}
Let $P$ be the final set returned by the {\spotg} method described in Algorithm~\ref{alg:SPOT-Greedy}. Let $\alpha = \alpha_{P,s}$ be the submodularity ratio of the set $P$ w.r.t. $s$. If $P^{\ast}$ is the optimal set of $k$ elements that maximizes $f(\cdot)$ in the SPOT optimization problem (\ref{eqn:prototype-selection}), then
\begin{equation}
f(P) \geq f\left(P^{\ast}\right) \left[ 1-e^{-\frac{1}{\alpha}} \right]  \geq f\left(P^{\ast}\right) \left[ 1-e^{-\frac{1}{s}} \right].
\end{equation}
When $s=1$ we recover the known approximation guarantee of $\left(1-e^{-1}\right)$~\cite{Nemhauser78}.
\end{theorem}

\begin{algorithm}[t]
    \caption{{\spotg}}
    \label{alg:SPOT-Greedy}
\begin{algorithmic}
\STATE \textbf{Input:} sparsity level $k$ or lower bound $\epsilon$ on increment in $f(.)$, $X$, $Y$, $s$, and $\bq$.
\STATE \textbf{Initialize} $P=\emptyset$
\WHILE{$|P|\le k$ or increment in objective $\ge \epsilon$.}
\STATE Define vector $\boldsymbol{\beta}$ with entries $\boldsymbol{\beta}_i = f\left(P \cup \{i\} \right) - f(P),$ $\forall i \in \left[m\right] \setminus P$.
\STATE {$S$} = Set of indices of top $s$ largest elements in $\boldsymbol{\beta}$. 
\STATE $P = P \cup S$.
\ENDWHILE
\STATE $\gamma_P=\argmax\limits_{\gamma\in\Gamma_P(\bq)} \inner{\bS_{P},\gamma}$; $\bwP=\gamma_P \bone$.
\STATE \textbf{Return} $P$, $\gamma_P$, $\bwP$. 
\end{algorithmic}
\end{algorithm}

\subsection{k-medoids as a special case of SPOT}

Consider the specific setting where the source and the target datasets are the same, i.e., $X=Y$. Let $n =|X|$ and $\bq_j = 1/n$ having uniform weights on the samples. Selecting a prototypical set $P \subset X$ is in fact a data summarization problem of choosing few representative exemplars from a given set of $n$ data points, and can be thought as an output of a clustering method where $P$ contains the the cluster centers. A popular clustering method is the k-medoids algorithm that ensures the cluster centers are exemplars chosen from actual data points~\cite{kmediods}. As shown in \cite{mirzasoleiman16a}, the objective function for the k-medoids problem is 
$$g(P) = \frac{1}{n}\sum\limits_{j=1}^n \maxop_{\bz \in P} l\left(\bz,\bx_j\right),$$ where $l\left(\bx_i,\bx_j\right) = \bS_{ij}$ 
defines the similarity between the respective data points. Comparing it against (\ref{eqn:reducedform}) gives a surprising connection that the \emph{k-medoids algorithm is a special case of learning an optimal transport plan with a sparse support in the setting where the source and target distributions are the same}. Though the relation between OT and k-means is discussed in~\cite{kmeansOT,cuturi14a}, we are not cognizant of any prior works that explains k-medoids from the lens of optimal transport. However, the notion of \emph{transport} loses its relevance as there is no distinct target distribution to which the source points need to be transported. It should be emphasized that the connection with k-medoids is only in the limited case where the source and target distributions are the same. Hence, the popular algorithms that solve the k-medoids problem~\cite{kmedoidsalgo} like PAM, CLARA, and CLARANS cannot be applied in the general setting when the distributions are {\it different}.

\section{Related works and discussion}\label{sec:related}

As discussed earlier, recent works \cite{Kim16,proto} view the unsupervised prototype selection problem as searching for a set $P\subset X$ whose underlying distribution is similar to the one corresponding to the target dataset $Y$. However, instead of the true source and target distributions, only samples from them are available. In such a setting, $\varphi$-divergences \cite{csiszar72a} e.g., the total variation distance and KL-divergence, among others require density estimation or space-partitioning/bias-correction techniques \cite{smola07a,song08a}, which can be computationally prohibitive in higher dimensions. Moreover, they may be agnostic to the natural geometry of the ground metric. 
The maximum mean discrepancy (MMD) metric (\ref{eqn:MMDObjective}) employed by \cite{Kim16,proto}, on the other hand, can be computed efficiently but does not faithfully lift the ground metric of the samples \cite{feydy18a}. 

We propose an optimal transport (OT) based prototype selection approach. OT framework respects the intrinsic geometry of the space of the distributions. Moreover, there is an additional flexibility in the choice of the ground metric, e.g., $\ell_1$-norm distance, which need not be a (universal) kernel induced function sans which the distribution approximation guarantees of MMD may no longer be applicable \cite{gretton12a}. Solving the classical OT problem (\ref{eqn:emd}) is known to be computationally more expensive than computing MMD. However, our setting differs from the classical OT setup, as the source distribution is also learned in (\ref{eqn:f}). As shown in Lemmas~\ref{lemma:welimination}\&~\ref{lemma:incrementalf}, the joint learning of the source distribution and the optimal transport plan has an equivalent but computationally efficient reformulation (\ref{eqn:fred}). 

Using OT is also favorable from a theoretical standpoint. Though the MMD function in \cite{Kim16} is proven to be submodular, it is only under restricted conditions like the choice of kernel matrix and equal weighting of prototypes. The work in \cite{proto} extends \cite{Kim16} by allowing for unequal weights and eliminating any additional conditions on the kernel, but forgoes submodularity as the resultant MMD objective (\ref{eqn:MMDObjective}) is only weakly submodular. In this backdrop, the SPOT objective function (\ref{eqn:prototype-selection}) is submodular without requiring any further assumptions. It is worth noting that submodularity leads to a tighter approximation guarantee of $\left(1-e^{-1}\right)$ using greedy approximation algorithms~\cite{Nemhauser78}, whereas the best greedy based approximation for weak submodular functions (submodularity ratio of $\alpha < 1$) is only $\left(1-e^{-\alpha}\right)$ \cite{weaksub}. A better theoretical approximation of the OT based subset selection encourages the selection of better quality prototypes.

\section{Experiments}\label{sec:experiment}
We evaluate the generalization performance and computational efficiency of the proposed approach against state-of-the-art on several real-world datasets. The codes are available at \url{https://pratikjawanpuria.com}. 

The following algorithms are evaluated. 
\begin{itemize}
    \item \textbf{MMD-Critic} \cite{Kim16}: it uses a maximum mean discrepancy (MMD) based scoring function. All the samples are weighted equally in the scoring function. 
    \item \textbf{ProtoDash} \cite{proto}: it uses a {\it weighted} MMD based scoring function. The learned weights indicate the importance of the samples.
    \item \textbf{\spots}: our fast heuristic algorithm described in Section~\ref{sec:SPOT-Simple}.
    \item \textbf{\spotg}: our greedy and incremental algorithm (Algorithm~\ref{alg:SPOT-Greedy}).
\end{itemize}

Following~\cite{sproto,Kim16,proto}, we validate of the quality of the representative samples selected by different prototype selection algorithms via the performance of the corresponding \textit{nearest prototype classifier}. 
Let $X$ and $Y$ represent source and target datasets containing different class distributions and let $P\subseteq X$ be a candidate representative set of the target $Y$. The quality of $P$ is evaluated by classifying the target set instances with $1$-nearest neighbour ($1$-NN) classifier parameterized by the elements in $P$. The class information of the samples in $P$ is made available during this evaluation stage. Such classifiers can achieve better generalization performance than the standard 1-NN classifier due to reduction of noise overfitting \cite{crammer02a} and have been found useful for large scale classification problems ~\cite{wohlhart13a,tibshirani02a}. 

\eat{Our implementation is available at \url{https://pratikjawanpuria.com}.}



\begin{figure*}[t]
\begin{center}
\begin{tabular}{cc}
\begin{minipage}{0.3\hsize}
\begin{center}
\includegraphics[width=\hsize]{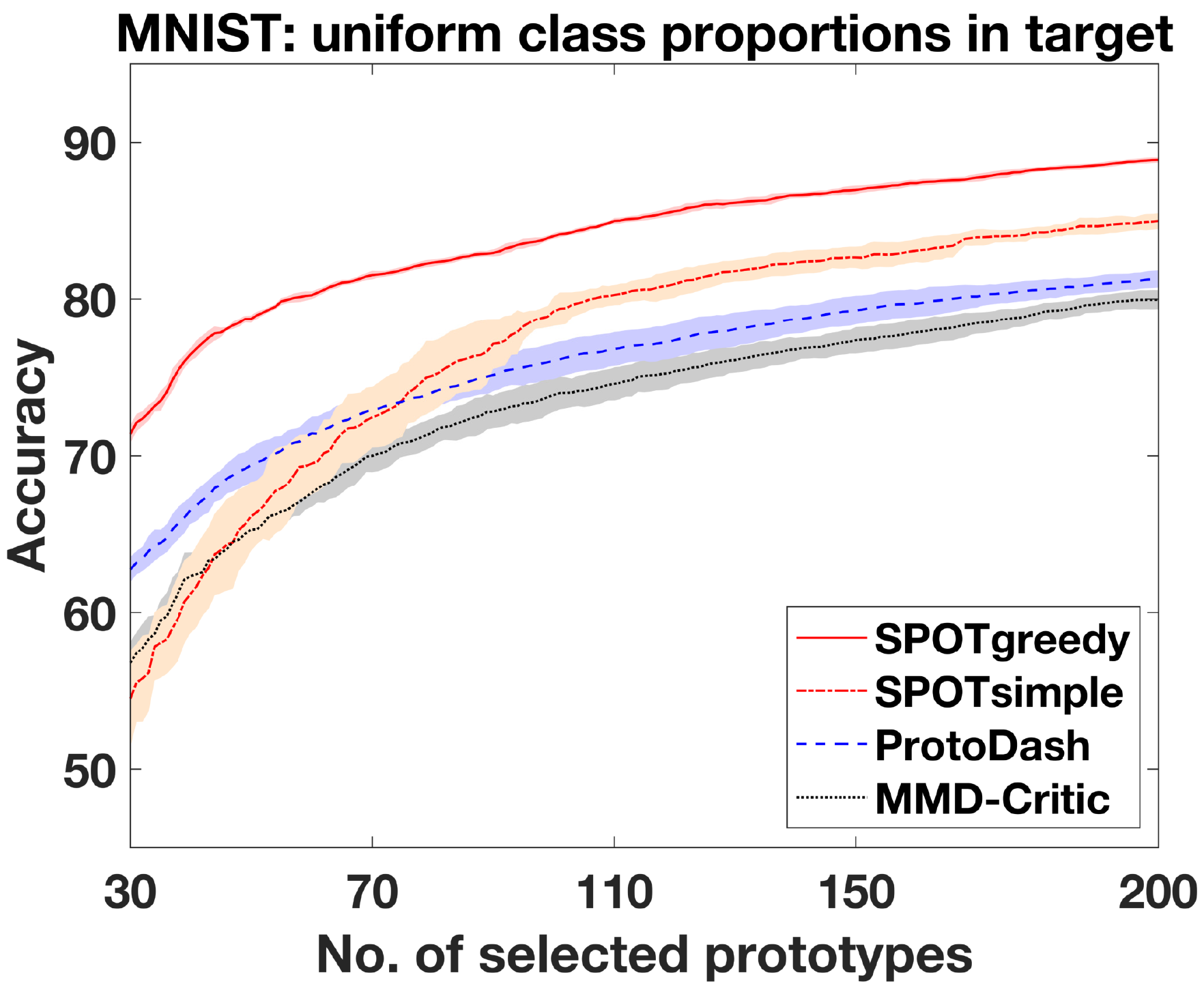}
\end{center}
\end{minipage}
\begin{minipage}{0.3\hsize}
\begin{center}
\includegraphics[width=\hsize]{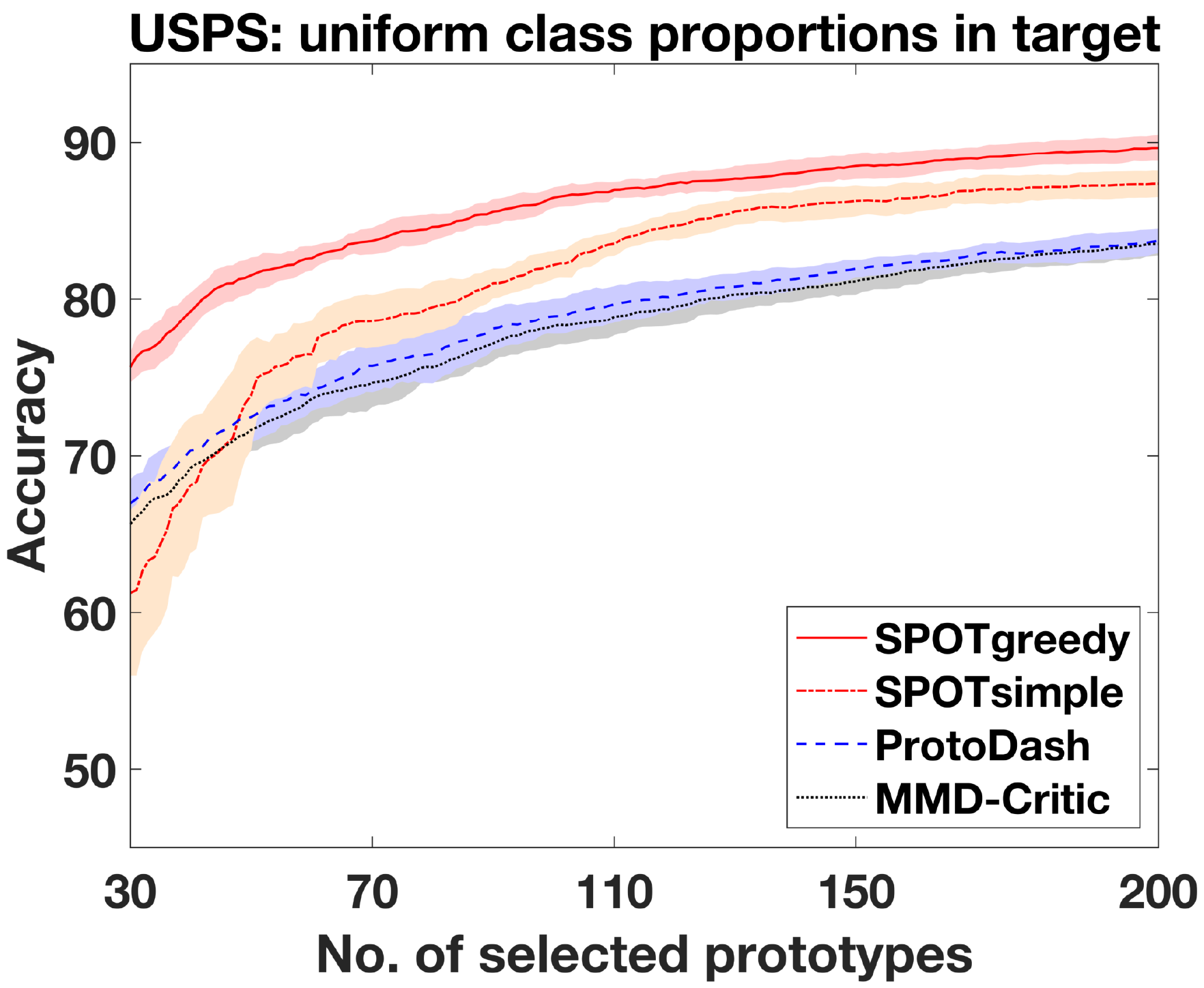}
\end{center}
\end{minipage}
\begin{minipage}{0.3\hsize}
\begin{center}
\includegraphics[width=\hsize]{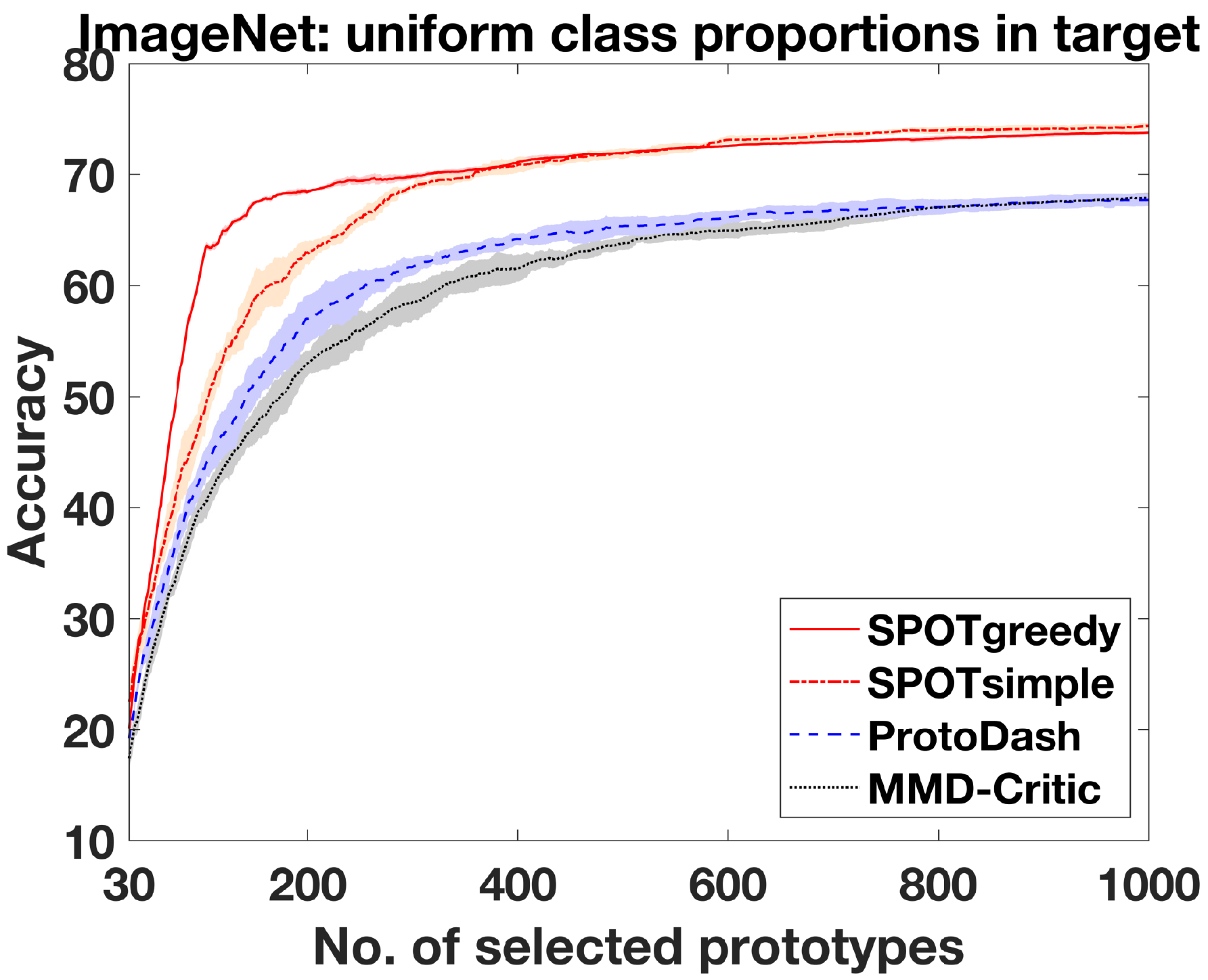}
\end{center}
\end{minipage}\\
\begin{minipage}{0.3\hsize}
\vspace{4pt}
\begin{center}
\includegraphics[width=\hsize]{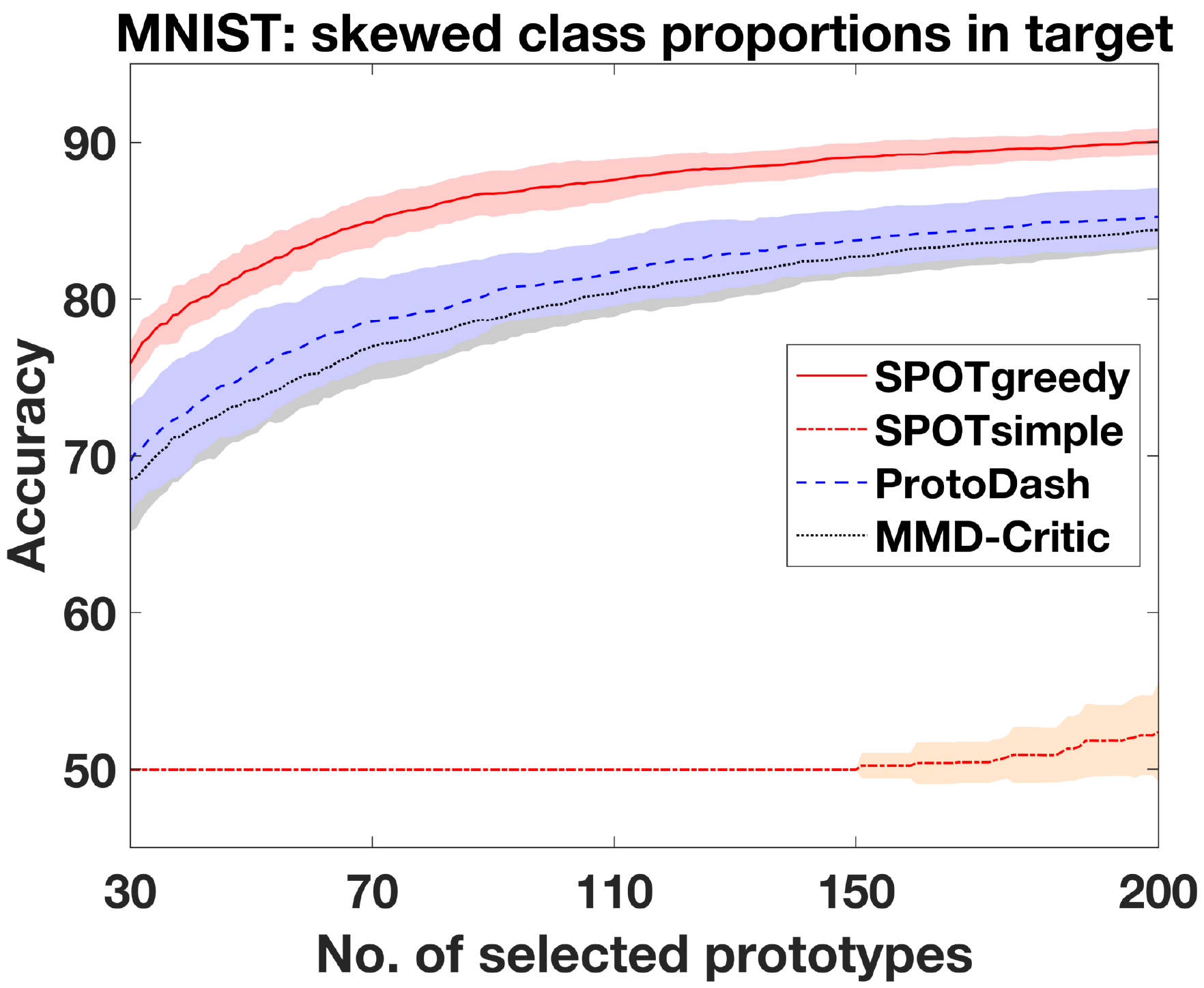}
\end{center}
\end{minipage}
\begin{minipage}{0.3\hsize}
\begin{center}
\includegraphics[width=\hsize]{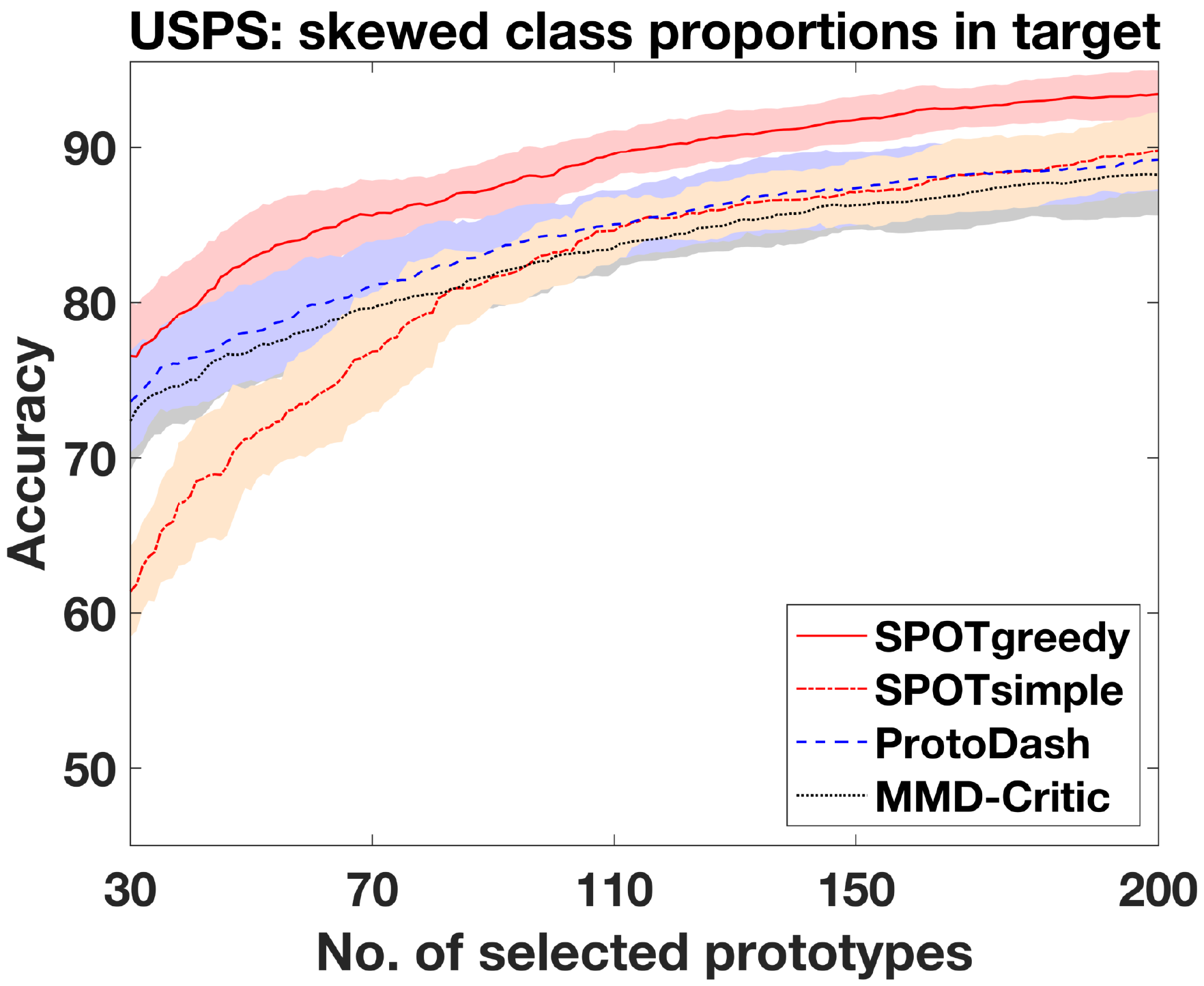}
\end{center}
\end{minipage}
\begin{minipage}{0.3\hsize}
\begin{center}
\includegraphics[width=\hsize]{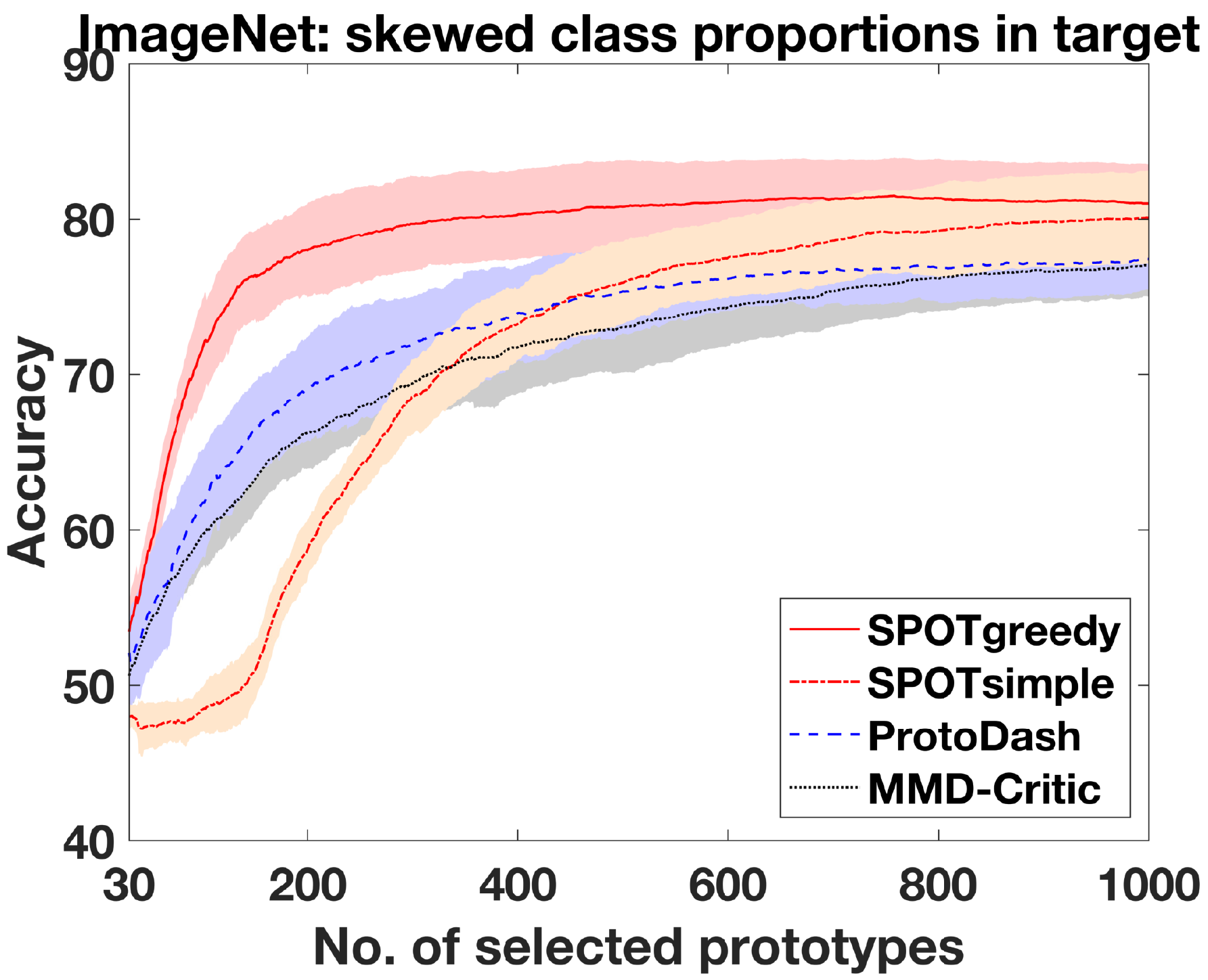}
\end{center}
\end{minipage}
\end{tabular}
\caption{
Performance of different prototype selection algorithms. The standard deviation for every $k$ is represented as a lighter shaded band around the mean curve corresponding to each method. 
[Top row] all the classes have uniform representation in the target set. [Bottom row] the challenging skewed setting where a randomly chosen class represents $50\%$ of the target set (while the remaining classes together uniformly represent the rest). 
}
\label{fig:sparsity}
\end{center}
\end{figure*}

\begin{figure*}[t]
\begin{center}
\begin{tabular}{ccc}
\begin{minipage}{0.3\hsize}
\begin{center}
\includegraphics[width=\hsize]{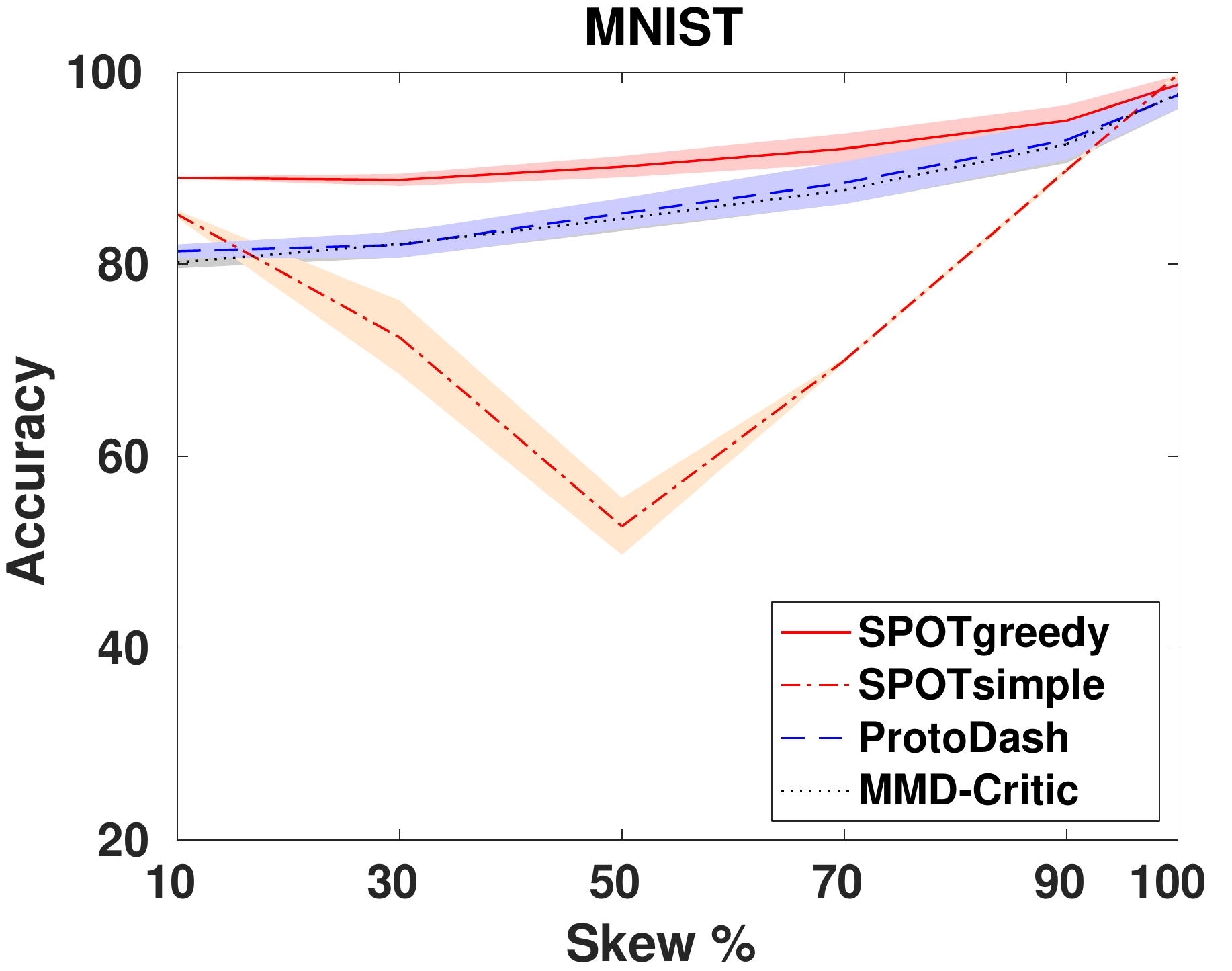}\\
{(a)}
\end{center}
\end{minipage}
&
\begin{minipage}{0.3\hsize}
\begin{center}
\includegraphics[width=\hsize]{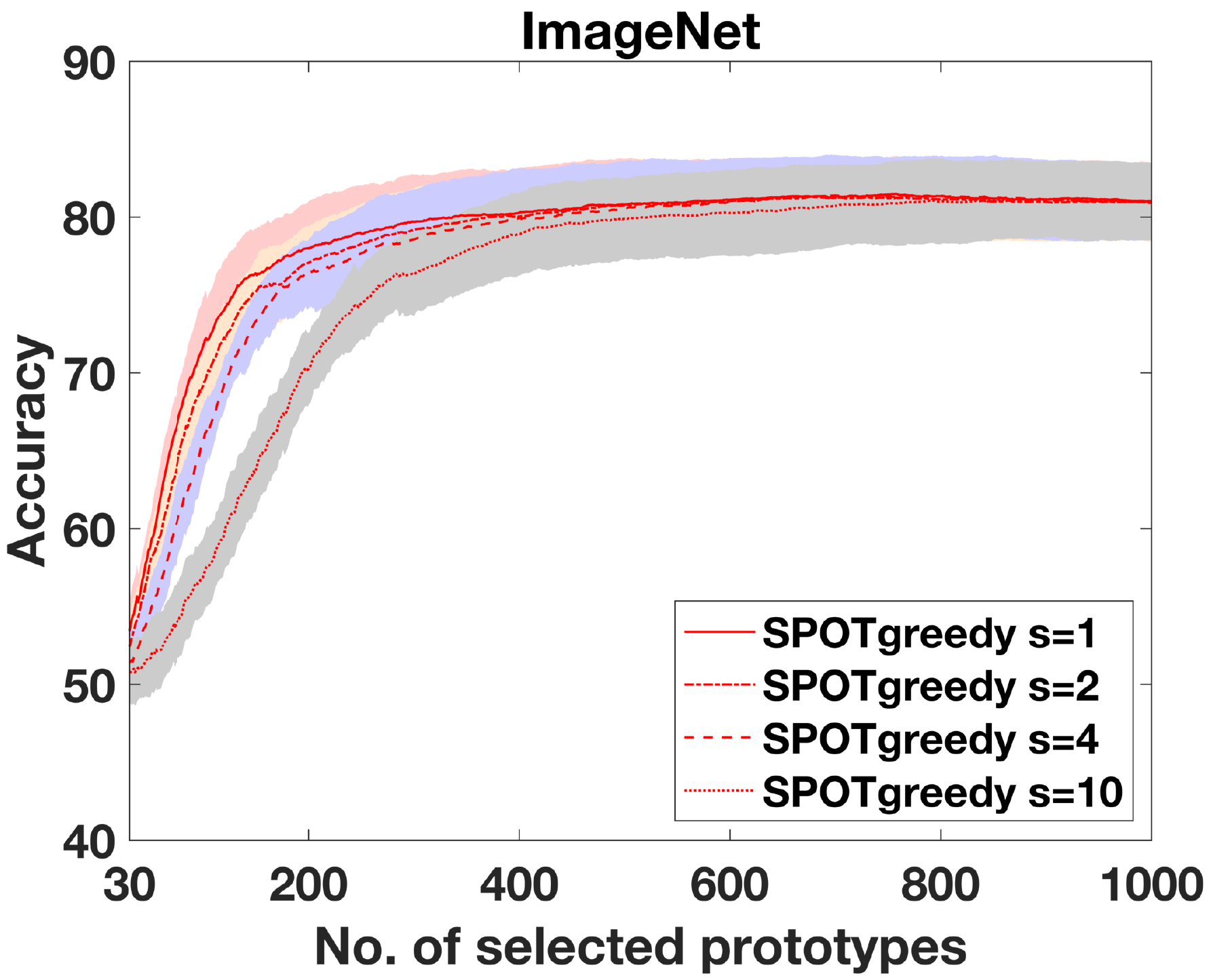}\\
{(b)}
\end{center}
\end{minipage}
&
\begin{minipage}{0.3\hsize}
\begin{center}
\includegraphics[width=\hsize]{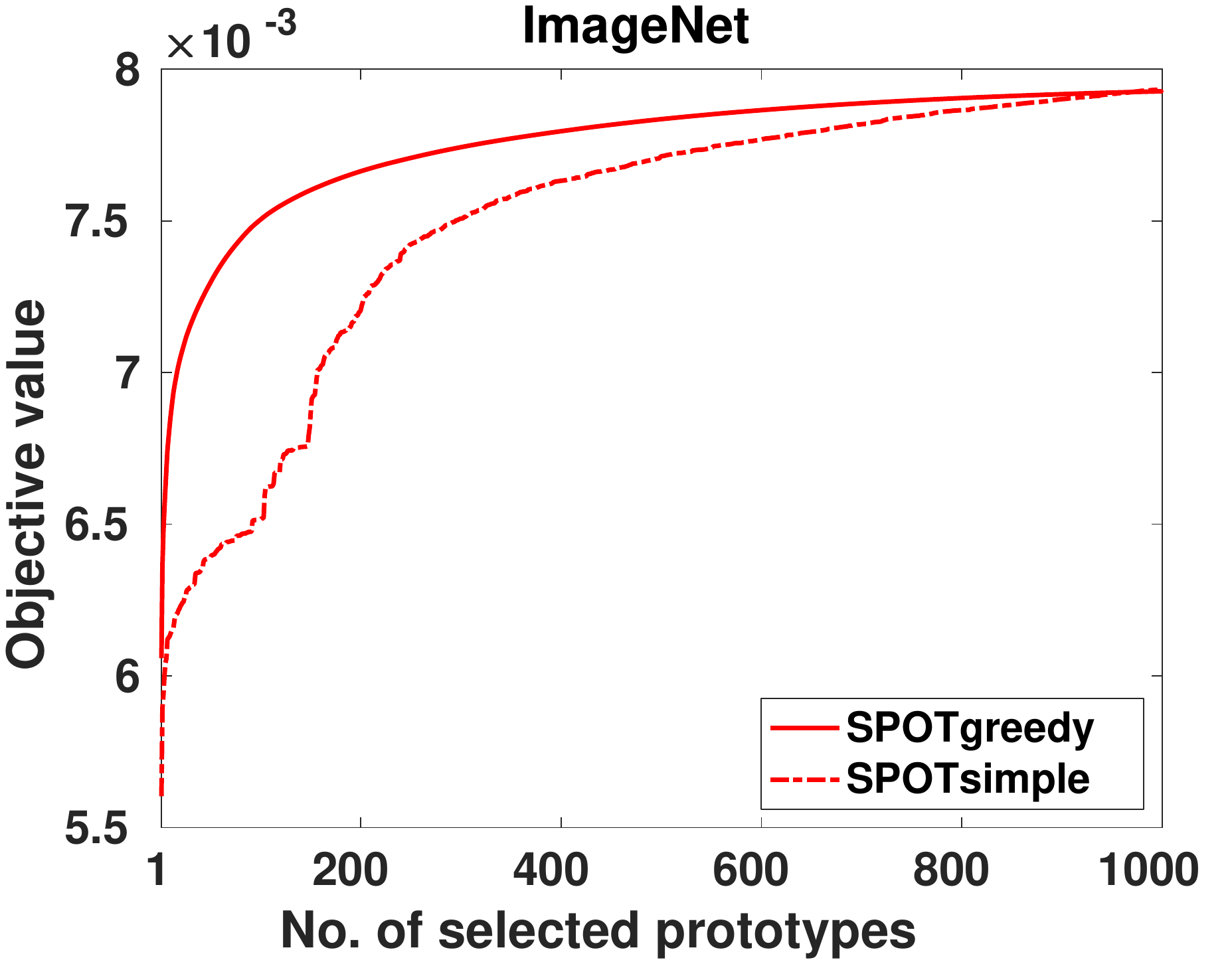}\\
{(c)}
\end{center}
\end{minipage}
\end{tabular}
\caption{(a) Comparisons of different algorithms in representing targets with varying skew percentage of a MNIST digit; (b) Performance of our {\spotg} algorithm with varying subset selection size $s$ on the ImageNet dataset; (c) Comparison of the objective value (\ref{eqn:prototype-selection}) obtained by the proposed algorithms {\spotg} and {\spots} for various values of $k$.
}
\label{fig:accuracy_vs_prop}
\end{center}
\end{figure*}

\subsection{Prototype selection within same domain}\label{subsec:exp-ps-same-domain}

We consider the following benchmark datasets. 
\begin{itemize}
    \item \textbf{ImageNet} \cite{imagenet15a}: we use the popular subset corresponding to ILSVRC 2012-2017 competition. The images have $2048$ dimensional deep features~\cite{he16a}. 
    \item \textbf{MNIST} \cite{lecun98a} is a handwritten digit dataset consisting of greyscale images of digits $\{0,\ldots,9\}$. The images are of $28 \times 28$ pixels. 
    \item \textbf{USPS} dataset \cite{hull94a} consists of handwritten greyscale images of $\{0,\ldots,9\}$ digits represented as $16\times16$ pixels.
    \item \textbf{Letter} dataset \cite{dua17a} consists of images of twenty-six capital letters of the English alphabets. Each letter is represented as a $16$ dimensional feature vector.
    \item \textbf{Flickr} \cite{thomee16a} is the Yahoo/Flickr Creative Commons multi-label dataset consisting of descriptive tags of various real-world outdoor/indoor images.
\end{itemize}
Results on the Letter and Flickr datasets are discussed in the appendix.

\noindent\textbf{Experimental setup.} In the first set of experiments, all the classes are equally represented in the target set. In second set of experiments, the target sets are skewed towards a randomly chosen class, whose instances (digit/letter) form $z\%$ of the target set and the instances from the other classes uniformly constitute the remaining $(100-z)\%$. For a given dataset, the source set is same for all the experiments and uniformly represents all the classes. Results are averaged over ten randomized runs. More details on the experimental set up are given in the appendix.\\ 


\noindent\textbf{Results.} Figure~\ref{fig:sparsity} (top row) shows the results of the first set of experiments on MNIST, USPS, and ImageNet. We plot the test set accuracy for a range of top-$k$ prototypes selected. We observe that the proposed {\spotg} outperforms ProtoDash and MMD-Critic over the whole range of $k$. 
Figure~\ref{fig:sparsity} (bottom row) shows the results when samples of a (randomly chosen) class constitutes $50\%$ of the target set. {\spotg} again {dominates} in this challenging setting. We observe that in several instances, {\spotg} opens up a significant performance gap even with only a few selected prototypes. The average running time on CPU of algorithms on the ImageNet dataset are: $55.0$s ({\spotg}), $0.06$s ({\spots}), $911.4$s (ProtoDash), and $710.5$s (MMD-Critic). We observe that both our algorithms, {\spotg} and {\spots}, are much faster than both ProtoDash and MMD-Critic. 

Figure \ref{fig:accuracy_vs_prop}(a) shows that {\spotg} achieves the best performance on different skewed versions of the MNIST dataset (with $k=200$). 
Interestingly, in cases where the target distribution is either uniform or heavily skewed, our heuristic non-incremental algorithm {\spots} can select prototypes that match the target distribution well. However, in the harder setting when skewness of class instances in the target dataset varies from $20\%$to $80\%$, {\spots} predominantly selects the skewed class leading to a poor performance. 

In Figure~\ref{fig:accuracy_vs_prop}(b), we plot the performance of {\spotg} for different choices of $s$ (which specifies the number of elements chosen simultaneously in each iteration). We consider the setting where the target has $50 \%$ skew of one of the ImageNet digits. Increasing $s$ proportionally decreases the computational time as the number of iterations $\left \lceil \frac{k}{s} \right \rceil$ steadily decreases with $s$. However, choosing few elements simultaneously generally leads to better target representation. 
We note that between $s=1$ and $s=10$, the degradation in quality is only marginal even when we choose as few as $110$ prototypes and the performance gap continuously narrows with more prototype selection. However, the time taken by {\spotg} with $s=10$ is $5.7$s, which is almost the expected $10$x speedup compared to {\spotg} with $s=1$ which takes $55.0$s. 
In this setting, we also compare the qualitative performance of the proposed algorithms in solving Problem~(\ref{eqn:prototype-selection}). Figure~\ref{fig:accuracy_vs_prop}(c) shows the objective value obtained after every selected prototype on ImageNet. {\spotg} consistently obtains a better objective than {\spots}, showing the benefit of the greedy and incremental selection approach.\\

\begin{figure}[t]
\begin{center}
\begin{tabular}{ccc}
\begin{minipage}{0.3\hsize}
\begin{center}
{%
\setlength{\fboxsep}{0pt}%
\setlength{\fboxrule}{0.46pt}%
\fbox{\includegraphics[width=\hsize]{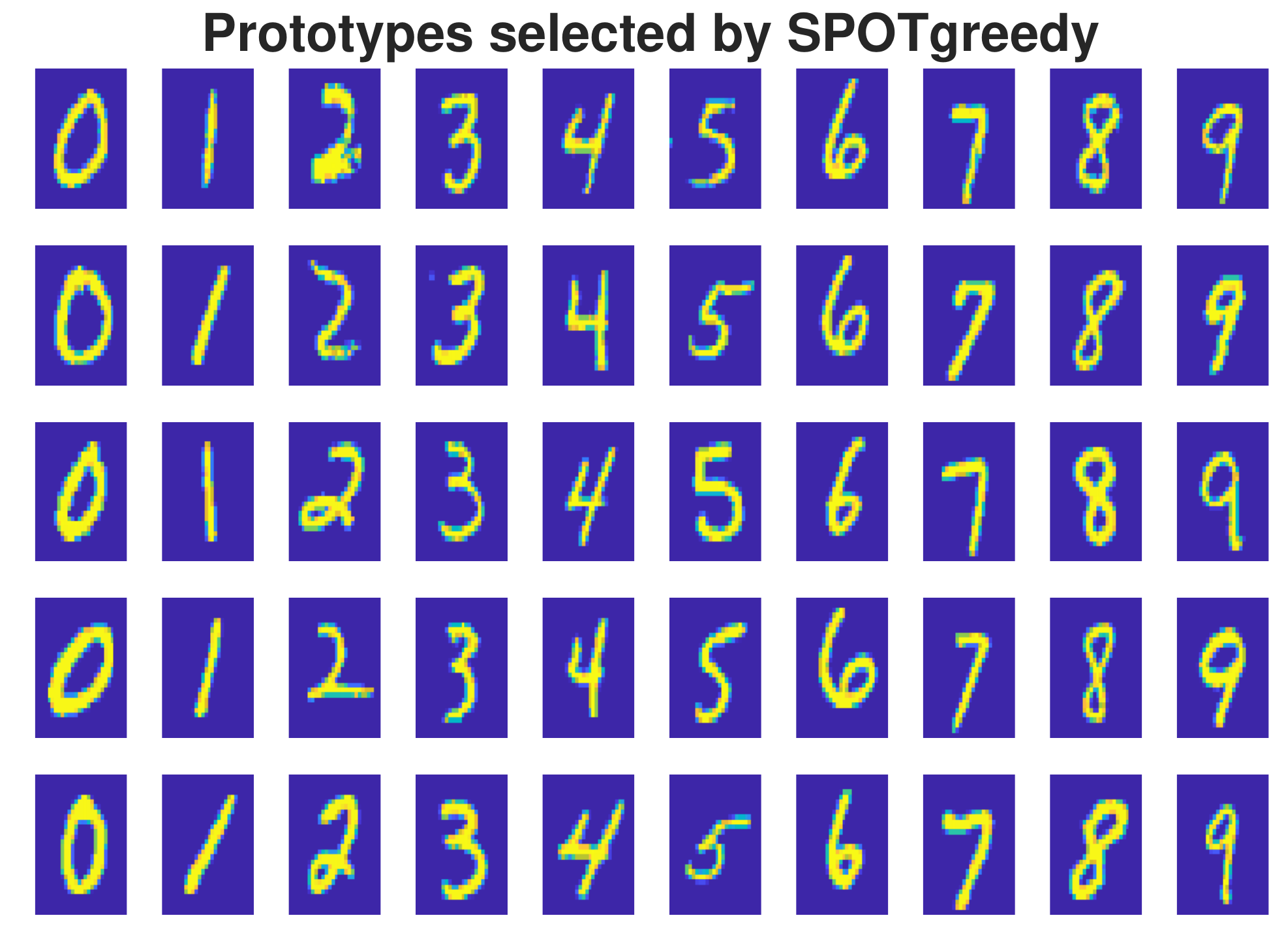}}\\
{(a)}
}%
\end{center}
\end{minipage}
& 
\begin{minipage}{0.3\hsize}
\begin{center}
{%
\setlength{\fboxsep}{0pt}%
\setlength{\fboxrule}{0.5pt}%
\fbox{\includegraphics[width=\hsize]{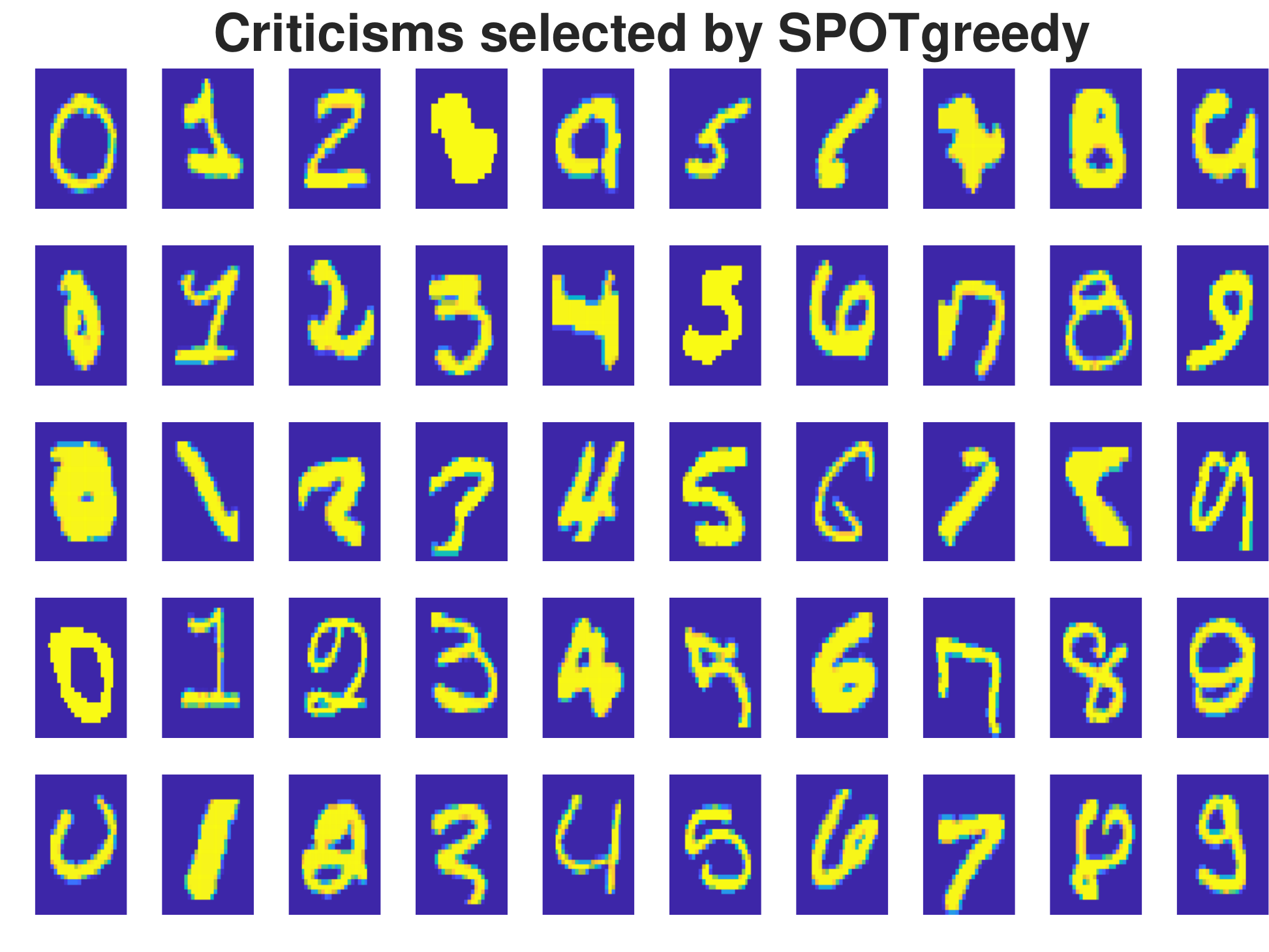}}\\
{(b)}
}%
\end{center}
\end{minipage}
&
\begin{minipage}{0.3\hsize}
\begin{center}
{%
\setlength{\fboxsep}{0pt}%
\setlength{\fboxrule}{0.5pt}%
\fbox{\includegraphics[width=\hsize]{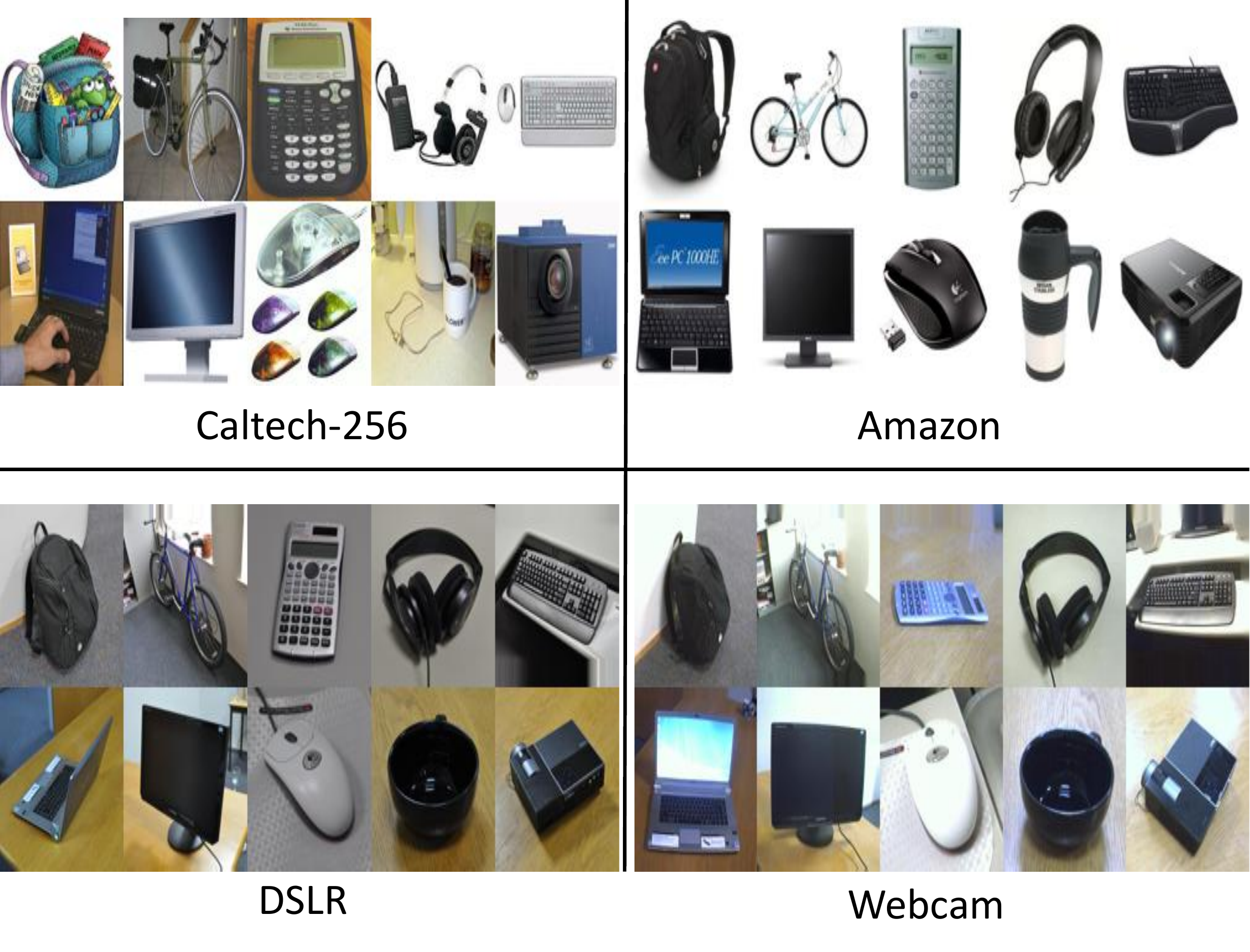}}\\
{(c)}
}%
\end{center}
\end{minipage}
\end{tabular}
\caption{(a) Prototypes selected by {\spotg} for the dataset containing one of the ten MNIST digits (column-wise); (b) Criticisms chosen by {\spotg} for the dataset containing one of the ten MNIST digits (column-wise); (c) Example images representing the ten classes in the four domains of the Office-Caltech dataset~\cite{gong12a}. }
\label{fig:MNISTPrototypesCriticisms}
\end{center}
\end{figure}

\noindent\textbf{Identifying criticisms for MNIST.} We further make use of the prototypes selected by {\spotg} to identify \emph{criticisms}. These are data points belonging to the region of input space not well explained by prototypes and are farthest away from them. We use a witness function similar to \cite[Section 3.2]{Kim16}.  
The columns of Figure~\ref{fig:MNISTPrototypesCriticisms}(b) visualizes the few chosen criticisms, one for each of the $10$ datasets containing samples of the respective MNIST digits. It is evident that the selected data points are indeed outliers for the corresponding digit class. Since the criticisms are those points that are maximally dissimilar from the prototypes, it is also a reflection on how well the prototypes of {\spotg} represent the underlying class as seen in Figure~\ref{fig:MNISTPrototypesCriticisms}(a), where in each column we plot the selected prototypes for a dataset comprising one of the ten digits. 

\begin{table*}[t]
\caption{Accuracy obtained on the Office-Caltech dataset.}\label{table:officeCaltech}
\centering
\begin{adjustbox}{max width=\textwidth}
{
\setlength{\tabcolsep}{4pt}
\begin{tabular}{lcccccc}
\toprule
{Task} & {MMD-Critic} & {{MMD-Critic+OT}} & {ProtoDash} & {ProtoDash+OT} &  {{\spots}} & {{\spotg}}\\
\midrule
$A\rightarrow C$ &   $73.98$ & $78.16$  & $70.23$ & $72.28$  & $82.62$ & $\mathbf{83.60}$\\
$A\rightarrow D$ &   $75.16$ & $72.61$  & $77.71$ & $71.97$  & $80.25$ & $\mathbf{82.80}$\\
$A\rightarrow W$ &   $51.53$ & $62.71$  & $48.81$ & $58.64$  & $62.37$ & $\mathbf{75.59}$\\
$C\rightarrow A$ &   $83.71$ & $86.17$  & $83.82$ & $87.25$  & $71.92$ & $\mathbf{90.03}$\\
$C\rightarrow D$ &   $70.06$ & $75.16$  & $71.34$ & $70.70$  & $75.80$ & $\mathbf{89.17}$\\
$C\rightarrow W$ &   $49.83$ & $54.92$  & $46.44$ & $53.56$  & $70.85$ & $\mathbf{82.03}$\\
$D\rightarrow A$ &   $82.85$ & $85.21$  & $83.39$ & $83.82$  & $\mathbf{91.00}$ & $90.89$\\
$D\rightarrow C$ &   $78.25$ & $78.34$  & $75.40$ & $79.41$  & $85.38$ & $\mathbf{86.27}$\\
$D\rightarrow W$ &   $80.00$ & $84.41$  & $85.08$ & $86.10$  & $75.59$ & $\mathbf{92.20}$\\
$W\rightarrow A$ &   $71.60$ & $78.56$  & $68.38$ & $74.71$  & $\mathbf{87.03}$ & $84.99$\\
$W\rightarrow C$ &   $67.20$ & $75.76$  & $65.86$ & $74.60$  & $74.06$ & $\mathbf{83.12}$\\
$W\rightarrow D$ &   $92.36$ & $\mathbf{96.18}$  & $88.54$ & $89.81$  & $86.62$ & $94.90$\\
\midrule
Average & $73.04$ & $77.35$  & $72.08$ & $75.24$ & $78.36$ & $\mathbf{86.30}$\\
\bottomrule
\end{tabular}
}
\end{adjustbox}
\end{table*}

\subsection{Prototype selection from different domains}\label{subsec:exp-ps-different-domain}
Section~\ref{subsec:exp-ps-same-domain} focused on settings where the source and the target datasets had similar/dissimilar class distributions. We next consider a setting where the source and target datasets additionally differ in feature distribution, e.g., due to covariate shift~\cite{candela09}.

Figure~\ref{fig:accuracy_vs_prop}(c) shows examples from the classes of the Office-Caltech dataset \cite{gong12a}, which has images from four domains: Amazon (online website), Caltech (image dataset), DSLR (images captured from a DSLR camera), and Webcam (images captured from a webcam). We observe that the images from the same class vary across the four domains  due to several factors such as different background, lighting conditions, etc. The number of data points in each domain is: $958$ (A: Amazon), $1123$ (C: Caltech), $157$ (D: DSLR), and $295$ (W: Webcam). The number of instances per class per domain ranges from $8$ to $151$. DeCAF6 features~\cite{donahue14a,courty17b} of size $4\,096$ are used for all the images. 
%
%
We design the experiment similar to Section~\ref{subsec:exp-ps-same-domain} by considering each domain, in turn, as the source or the target. There are twelve different tasks where task $A\rightarrow W$ implies that Amazon and Webcam are the source and the target domains, respectively. 

\textbf{Results.} Table~\ref{table:officeCaltech} reports the accuracy obtained on every task. We observe that our {\spotg} significantly outperforms MMD-Critic and ProtoDash. This is because {\spotg} learns both the prototypes as well as the transport plan between the prototypes and the target set. The transport plan allows the prototypes to be transported to the target domain via the barycentric mapping, a characteristic of the optimal transport framework. {\spotg} is also much better than {\spots} due to its superior incremental nature of prototype selection. 
We also empower the non-OT based baselines for the domain adaptation setting as follows. After selecting the prototypes via a baseline, we learn an OT plan between the selected prototypes and the target data points by solving the OT problem (\ref{eqn:emd}). The distribution of the prototypes is taken to be the normalized weights obtained by the baseline. This ensures that the prototypes selected by  MMD-Critic+OT, and ProtoDash+OT are also transported to the target domain. Though we observe marked improvements in the performance of MMD-Critic+OT and ProtoDash+OT, the proposed {\spotg} and {\spots} still outperform them. 

\section{Conclusion}\label{sec:conclusion}
We have looked at the prototype selection problem from the viewpoint of optimal transport. In particular, we show that the problem is equivalent to learning a sparse source distribution $w$, whose probability values $\bw_i$ specify the relevance of the corresponding prototype in representing the given target set.
After establishing connections with submodularity, we proposed the {\spotg} algorithm that employs incremental greedy selection of prototypes and comes with (i) deterministic theoretical guarantees, (ii) simple implementation with updates that are amenable to parallelization, and (iii) excellent performance on different benchmarks. 



\textbf{Future works}: We list a few interesting generalizations and research directions worth pursuing.
\begin{itemize}[noitemsep,topsep=0pt]
    \item The proposed $k$-prototype selection problem (\ref{eqn:prototype-selection}) may be viewed as learning a $\ell_0$-norm regularized (fixed-support) Wasserstein barycenter of a single distribution. Extending it to learning sparse Waserstein barycenter of multiple distributions may be useful in applications like model compression, noise removal, etc. 
    \item With the Gromov-Wasserstein (GW) distance \cite{memoli11a,peyre16a}, the OT distance has been extended to settings where the source and the target distributions do not share the same feature and metric space. 
    Extending SPOT with the GW-distances is useful when the source and the target domains share similar concepts/categories/classes but are defined over different feature spaces. 
\end{itemize}

%% file: supplementary.tex
\appendix


\section{Proofs}
\subsection{Proof of Lemma~\ref{lemma:welimination}}
Since $supp(\bw) \subseteq P$, (\ref{eqn:f}) can be equivalently stated as:
\begin{equation}
\label{eq:equivalentf}
f(P) \coloneqq \maxop_{\bw}\  \maxop_{\gamma\in\Gamma(\bw,\bq)} \inner{\bS_P,\gamma},
\end{equation}
where the optimization for the transport plan $\gamma$ is over dimensions $k \times n$ and $\bw$ is of length $k$. Let $\left(\bw^f,\gamma^f\right)$ be the point of maximum for $f(P)$. For the function $g(P) = \maxop_{\gamma\in\Gamma_P(\bq)} \inner{\bS_{P},\gamma}$, let the maximum occur at $\gamma^g$. 

Define $\bw^g = \gamma^g \bone$.  Observe that $\|\bw^g\|_1 = \sum\limits_{i=1}^m \sum\limits_{j=1}^n \gamma^g_{i,j} =  \sum\limits_{j=1}^n \bq_j = 1.$ Further as $\bw^g_i \geq 0$ and $supp\left(\bw^g\right) \subseteq P$, it is feasible source distribution in the optimization for $f(P)$. Assume $\gamma^f \neq \gamma^g$. We consider three different cases.\\
\noindent \textbf{case 1:} Let $f(P) = g(P)$. Then $\left(\bw^g,\gamma^g\right)$ also maximizes $f(P)$ proving that both the optimization problems are equivalent. \\
\noindent \textbf{case 2:} Let $f(P) < g(P)$. Then $\left(\bw^f,\gamma^f\right)$ cannot be the point of maximum as the value of the objective $\inner{\bS_P,\gamma^g}$ in~(\ref{eq:equivalentf}), evaluated at the feasible point $\left(\bw^g,\gamma^g\right)$, is higher than $\inner{\bS_P,\gamma^f}$. \\
\noindent \textbf{case 3:} Let $f(P) > g(P)$. Then $\gamma^g$ cannot be the maximum point for $g(P)$ as it can be further maximized by selecting the transport plan $\gamma^f$.\\
Hence $f(P) = g(P)$ and the proof follows.
\subsection{Proof of Lemma~\ref{lemma:incrementalf}}
Letting $\kappa^j_P$ be the maximum value in the vector $\bS^j_P$, define a function $f^j(P)$ as
\begin{equation}
\label{def:fjP}
    f^j(P) \coloneqq \bq_j \kappa^j_P
\end{equation}
so that $f(P) = \sum\limits_{j=1}^n f^j(P)$ from (\ref{eqn:reducedform}). Note that $\kappa^j_{P \cup S} = \maxop{\left(\kappa^j_P, \kappa^j_S\right)}$. Computing $\kappa^j_S$ is an $O(s)$ operation requiring to identify the maximum value of $s$ elements. Given $f(P)$ $\left(\mbox{and } \kappa^j_P, \forall j\right)$, $\kappa^j_{P \cup S}$ for each $j$ can be computed independently of each other in $O(s)$ and the lemma follows.

\subsection{Proof of Lemma~\ref{thm:submodular}}
The proof follows along similar lines as showing the k-medoids objective is submodular~\cite{mirzasoleiman16a}. We present the proof here for completeness. Consider the definition of $f^j(P)$ in (\ref{def:fjP}). As sums of monotone and submodular functions are also respectively monotone and submodular~\cite{fujishige05}, it is sufficient to prove that $f^j(P)$ inherits these characteristics.

 Consider any two sets $A \subseteq B$. As $\kappa^j_A \leq \kappa^j_B$, we have $f^j(A) \leq f^j(B)$ proving that it is monotone. For any $i \notin B$, let $\hat{A} = A \cup \{i\}$ and $\hat{B} = B \cup \{i\}$. If $\kappa^j_{\hat{B}} > \kappa^j_B$, then the maximum value in the $j^{th}$ column vector \emph{strictly increases} by adding the element $i$. Hence $\kappa^j_{\hat{A}} = \kappa^j_{\hat{B}}$. It then follows that $\kappa^j_{\hat{A}}-\kappa^j_A \geq \kappa^j_{\hat{B}}-\kappa^j_B$ proving that it is submodular.

\subsection{Proof of Theorem~\ref{thm:bounds}}
Let $t=\frac{k}{s}$ be the total number of iterations executed by {\spotg}. Without loss of generality we assume $s$ divides $k$. Denote $P_i$ as the set chosen at the end of iteration $i$ such that the final set $P=P_t$. Let $P_{i+1} = P_{i} \cup S_{i+1}$ created by adding the $s$ new elements in $S_{i+1}$ to $P_i$ during the iteration $i+1$. Define the residual set $P_R = P^{\ast} \backslash P_i$. Since $S_{i+1}$ contains the top $s$ elements that results in the maximum incremental gain, we have
\begin{equation*}
\frac{\sum\limits_{e \in S_{i+1}} \left[f\left(P_i \cup \{e\}\right) - f\left(P_i\right)\right]}{s} \geq \frac{\sum\limits_{e \in P_R} \left[ f\left(P_i \cup \{e\}\right) - f\left(P_i\right)\right]}{k},
\end{equation*}
where we have used the fact that $\left|P_R\right| \leq k$. Based on the definition of submodularity ratio in (\ref{eqn:subratio}) and $\alpha = \alpha_{P,s}$ in (\ref{eqn:subratioPs}), and recalling that $P_i \subseteq P$, we get
\begin{equation}\label{eq:PiPR}
\begin{array}{lll}
f\left(P_{i+1}\right) - f\left(P_i\right) &\geq& \frac{1}{\alpha}\sum\limits_{e \in S_{i+1}} \left[f\left(P_i \cup \{e\}\right) - f\left(P_i\right)\right]\\ 
&\geq& \frac{1}{\alpha} \frac{s}{k} \left[f\left(P_i \cup P_R\right) - f\left(P_i\right)\right].
\end{array}
\end{equation}
The last inequality in (\ref{eq:PiPR}) follows from the fact that submodularity ratio of $f(.)$ for the ordered pair $\left(P_i, P_R\right)$ is lower bounded by $1$. As $f(.)$ is monotone and $P^{\ast} \subseteq P_i \cup P_R$, we get $f\left(P^{\ast}\right) \leq f\left(P_i \cup P_R\right)$. Setting $\beta_i = f\left(P^{\ast}\right) - f\left(P_i\right)$ we can express $f\left(P_{i+1}\right) - f\left(P_i\right) = \beta_i - \beta_{i+1}$. Putting all this together and letting $\rho = \frac{s}{k} \frac{1}{\alpha}$, the increment at the iteration $i+1$ respects the inequality $\beta_i - \beta_{i+1} \geq \rho \beta_i$,
leading to the recurrence relation: $\beta_{i+1} \leq \left(1-\rho \right) \beta_i$. When iterated $t$ times from step $0$ and noting that $\beta_0 =  f\left(P^{\ast}\right)$ and $\beta_t = f\left(P^{\ast}\right) - f(P)$, we have
\begin{equation*}
     f(P) \geq f\left(P^{\ast}\right) \left[1-\left(1-\rho\right)^t\right].
\end{equation*}
Using the relation $1-\rho \leq e^{-\rho}$ for all $\rho \geq 0$ we have the required approximation guarantee:
\begin{equation*}
   f(P) \geq f\left(P^{\ast}\right) \left[1-e^{-\rho t}\right] \geq f\left(P^{\ast}\right) \left[1-e^{-\frac{1}{\alpha}}\right].
\end{equation*}

\begin{figure*}[t]
\begin{center}
\begin{tabular}{cc}
\begin{minipage}{0.46\hsize}
\begin{center}
\includegraphics[width=\hsize]{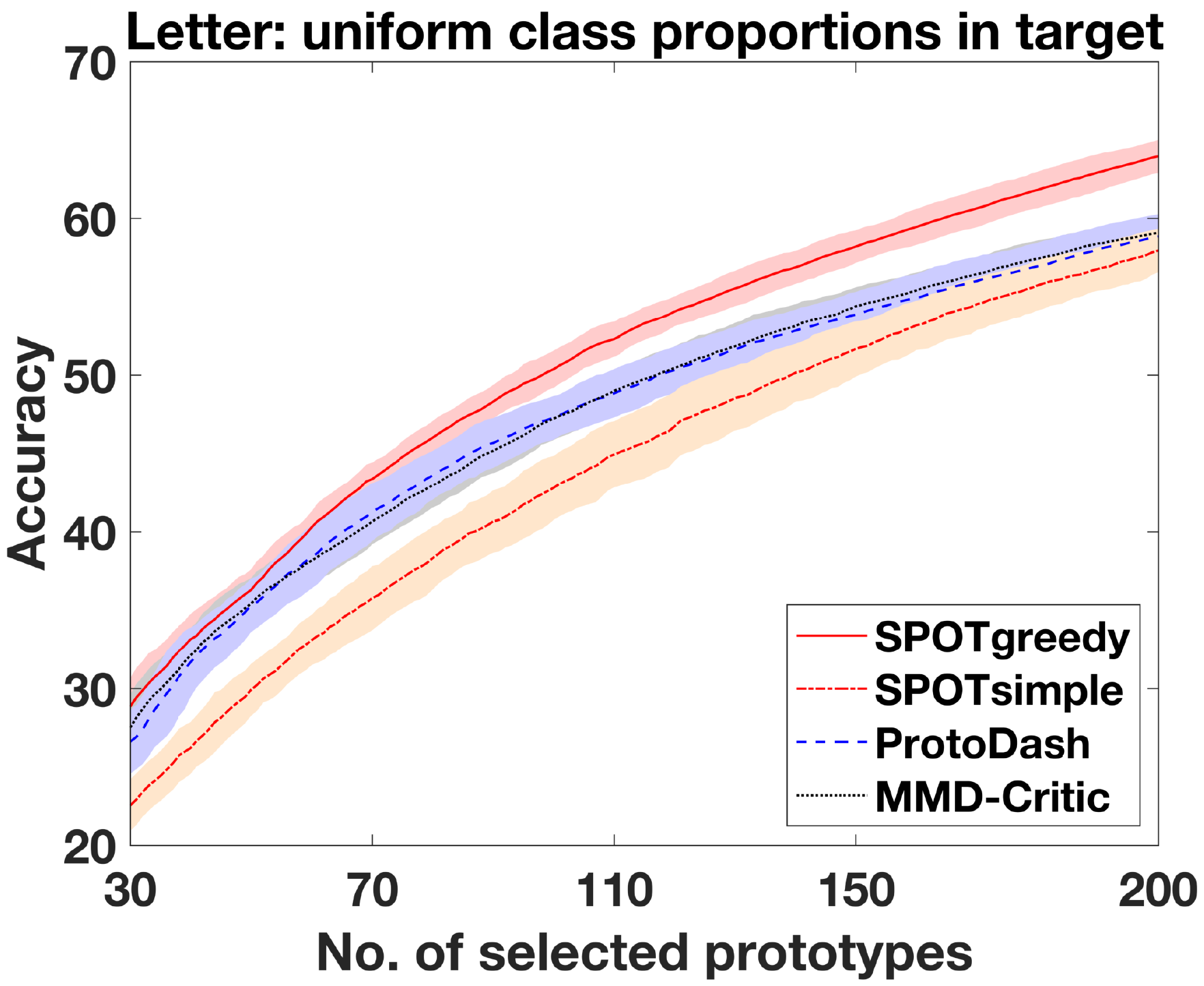}
\end{center}
\end{minipage}
\begin{minipage}{0.46\hsize}
\begin{center}
\includegraphics[width=\hsize]{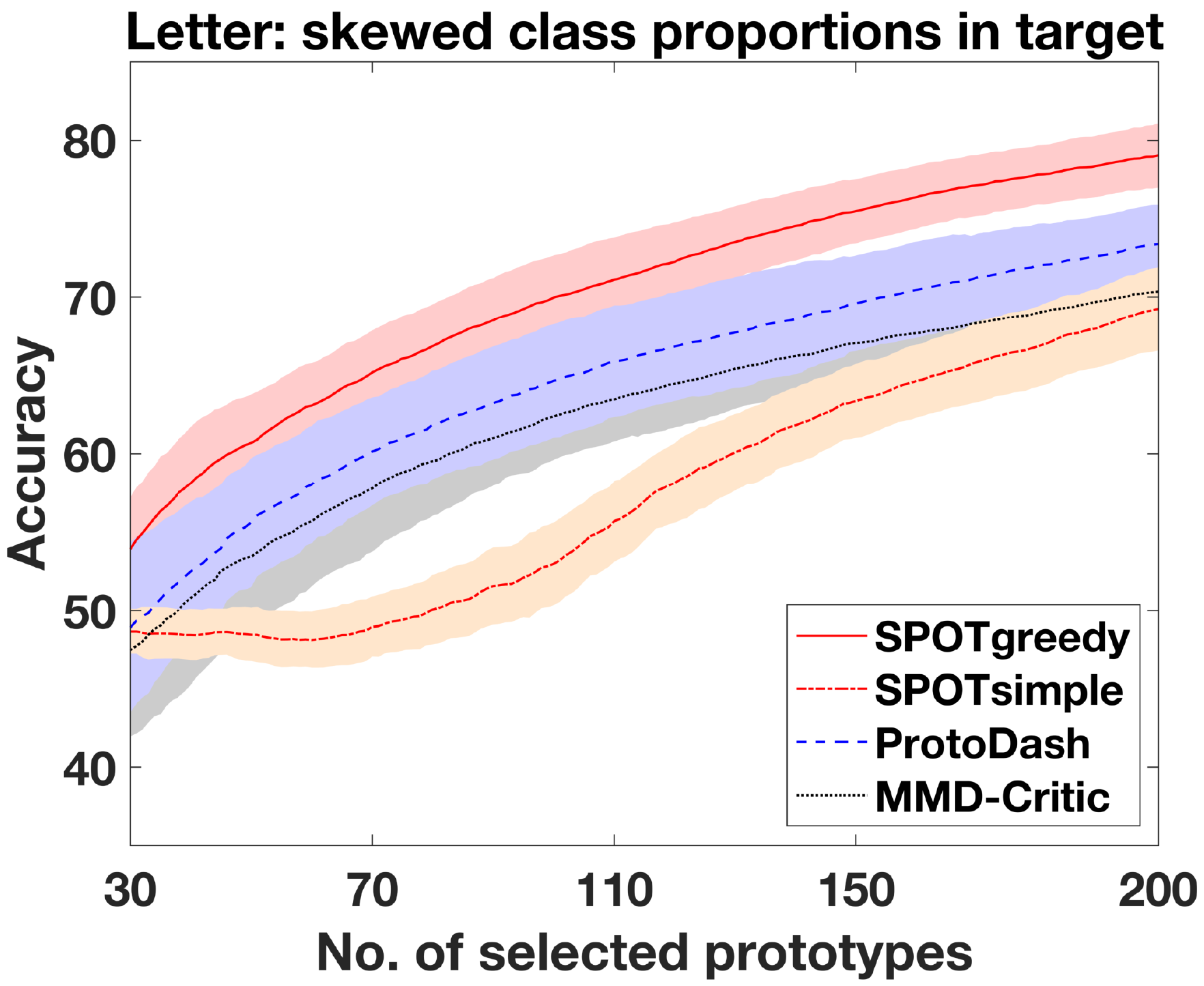}
\end{center}
\end{minipage}
\end{tabular}
\caption{Performance of different algorithms for prototype selection on the Letter dataset. In the challenging skewed setting, a randomly chosen class represents $50\%$ of the target set and the other $25$  classes are represented uniformly. 
}
\label{fig:letter_plots}
\end{center}
\end{figure*}

\begin{figure}[t]
\begin{center}
\begin{tabular}{cc}
\begin{minipage}{0.33\hsize}
\begin{center}
\includegraphics[width=\hsize]{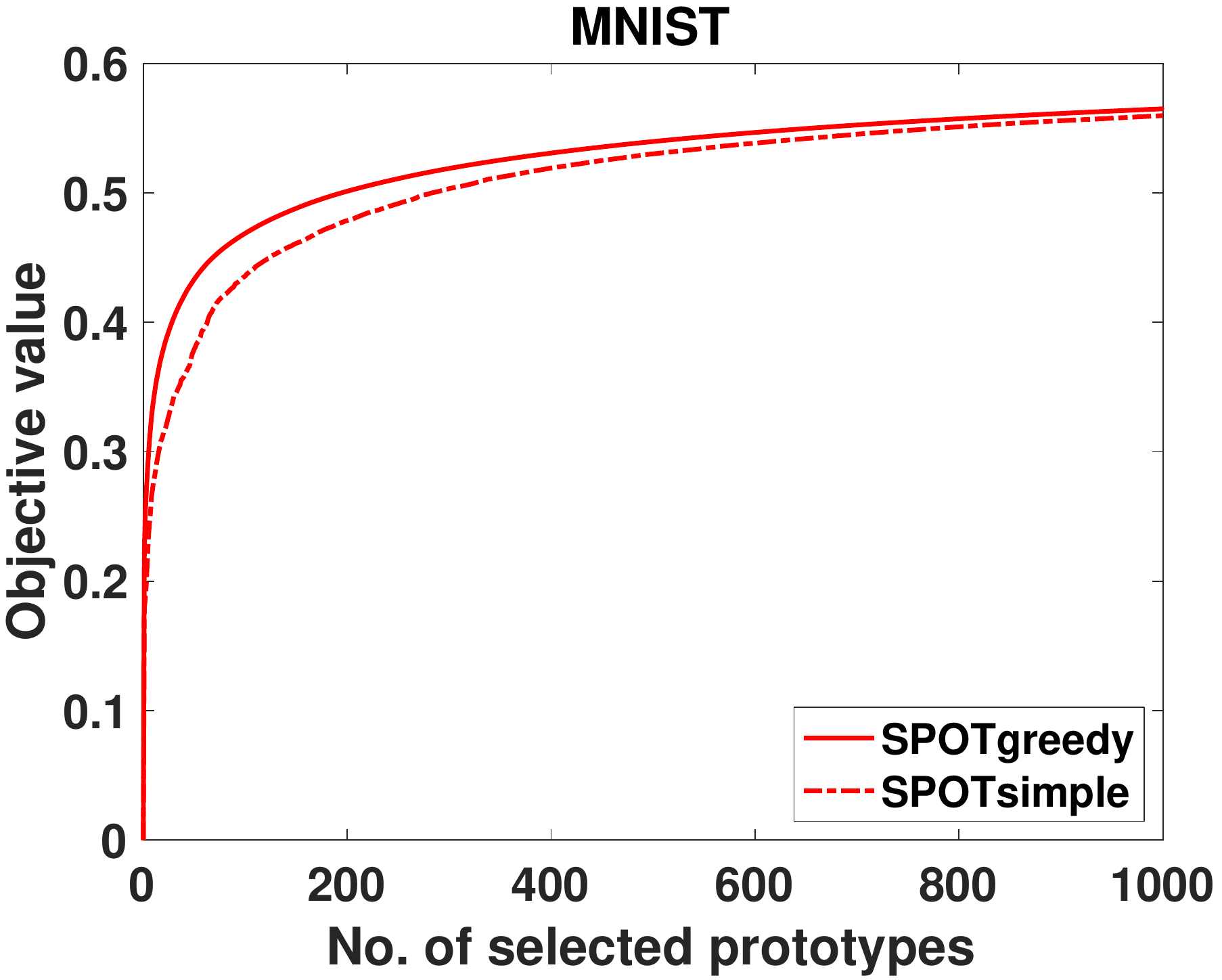}\\
{(a) The target set has uniform class proportions.}
\end{center}
\end{minipage}
\begin{minipage}{0.33\hsize}
\begin{center}
\includegraphics[width=\hsize]{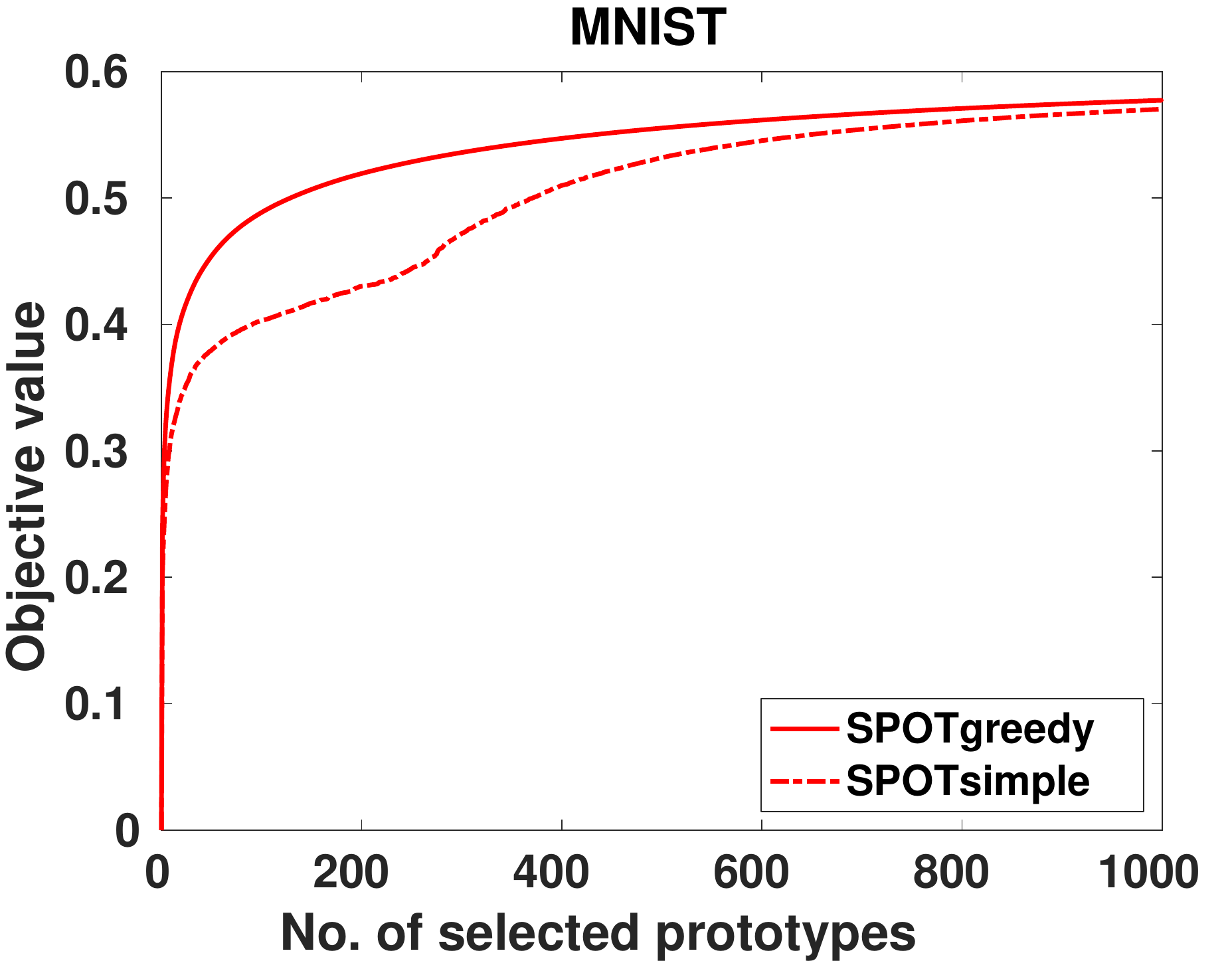}\\
{(b) The target set has skewed ($50\%$) class proportions.}
\end{center}
\end{minipage}
\begin{minipage}{0.33\hsize}
\begin{center}
\includegraphics[width=\hsize]{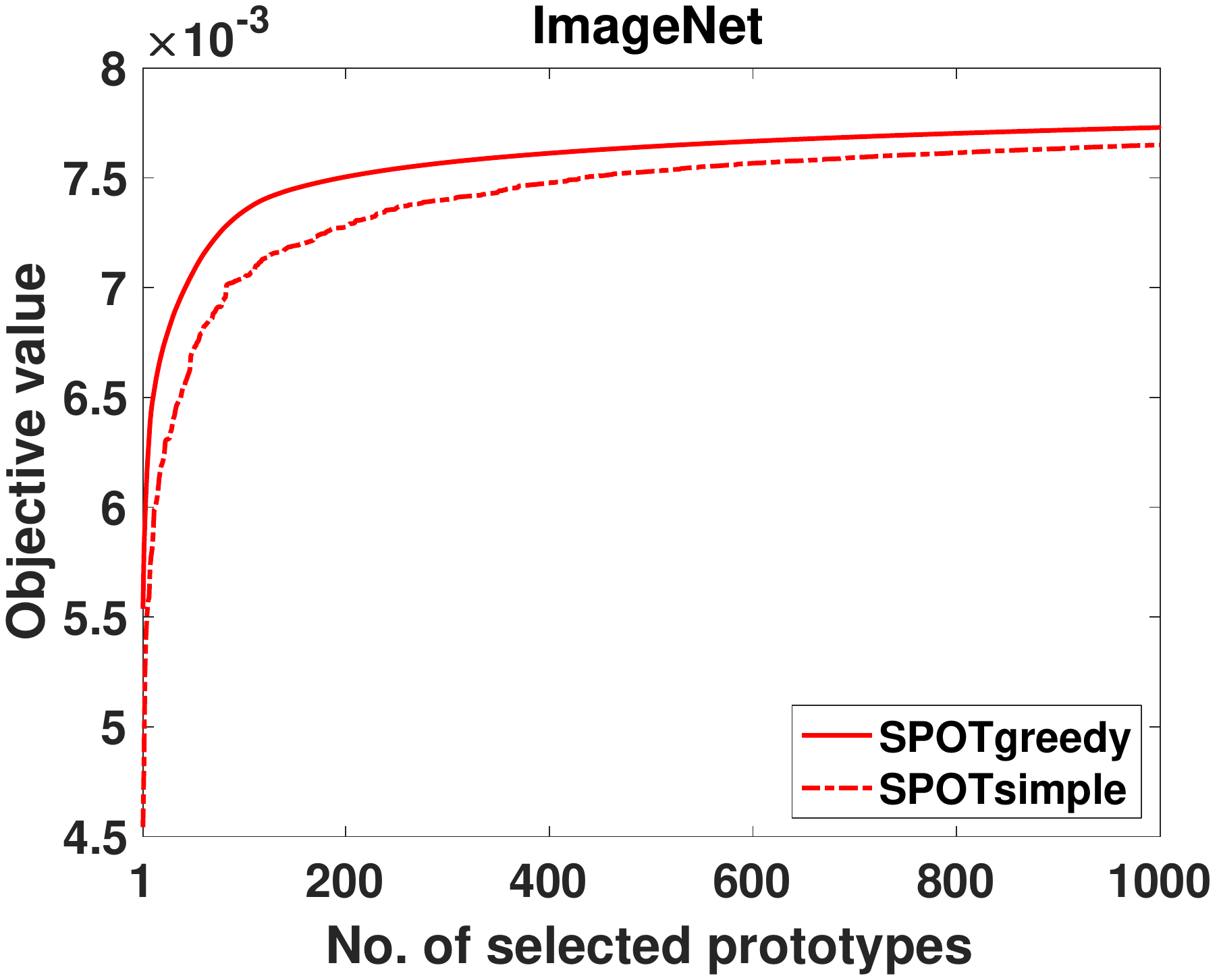}\\
{(c) The target set has skewed ($50\%$) class proportions.}
\end{center}
\end{minipage}
\end{tabular}
\caption{Performance of {\spots} and {\spotg} on the MNIST and ImageNet datasets. {\spotg} consistently obtains a better objective value.}
\label{fig:objective_mnist}
\end{center}
\end{figure}

\begin{figure}
\begin{center}
\begin{tabular}{cc}
\begin{minipage}{0.7\hsize}
\begin{center}
\includegraphics[width=\hsize]{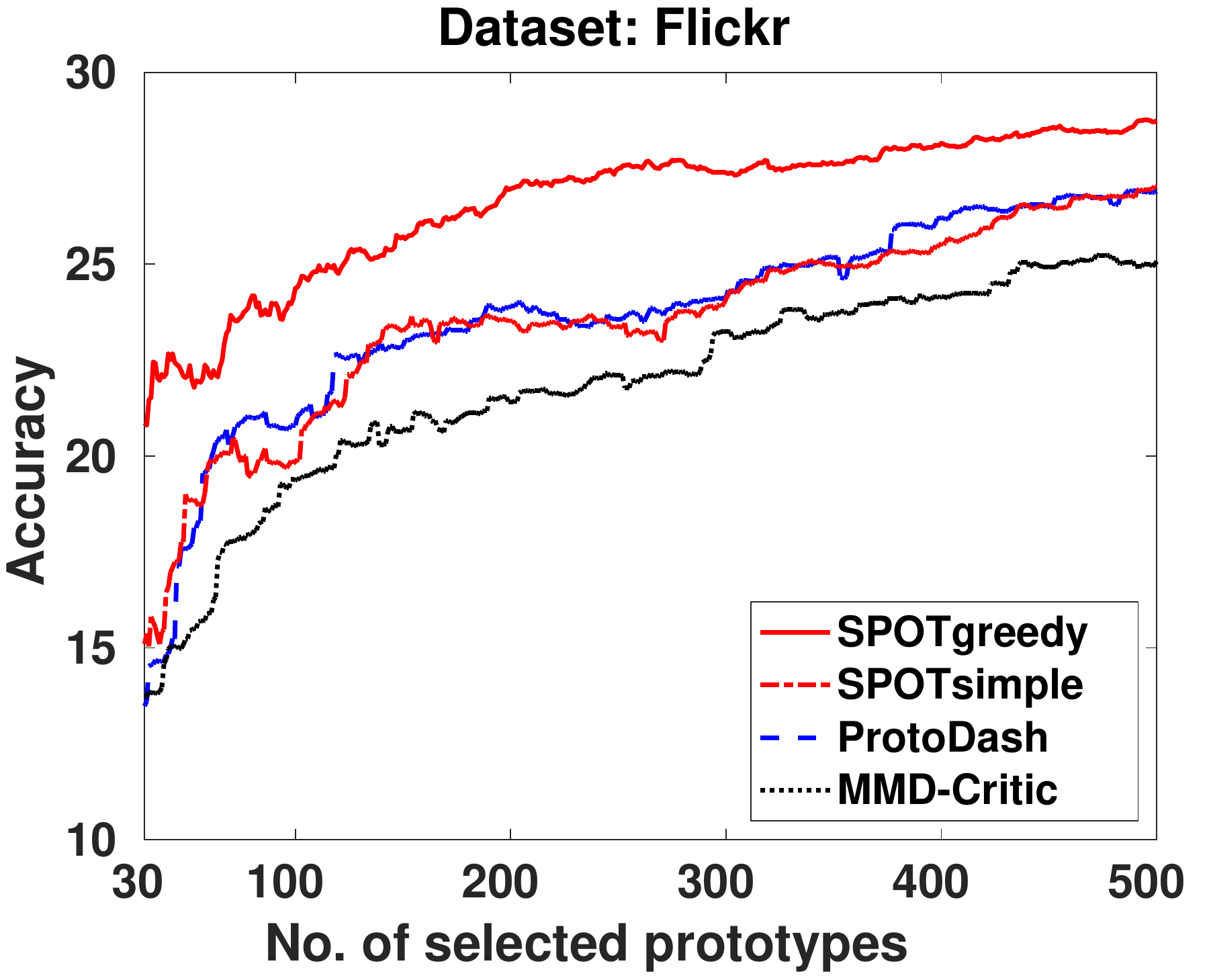}
\end{center}
\end{minipage}
\end{tabular}
\caption{Performance of different algorithms on the Flickr dataset.}
\label{fig:flickr}
\end{center}
\end{figure}

\section{Datasets and baselines details}
In this section, we present the details such as size of the source/target datasets and cross-validation on the hyper-parameters of the baselines. We begin with the dataset details:
\begin{itemize}
    \item \textbf{MNIST}\footnote{\url{http://yann.lecun.com/exdb/mnist}.}:
    It consists of two different sets of sizes $60K$ and $10K$ respectively. Following \cite{proto}, we randomly sampled $5K$ points from the $10K$ set and created the source set $X$. This source set is kept unchanged for all the (MNIST) experiments. The target set $Y$, constructed as a subset of $60K$, 
    varies with the skew of the randomly chosen class $c$. The instances from $c$ form $z=\{10,30,50,70,100\}$ percent of $Y$ and the instances from other classes uniformly constitute the remaining $(100-z)\%$ of $Y$. 
    The most frequent class in the MNIST training set has $6742$ elements while the least frequent class has $5421$ instances. Hence, when $z=10$, $Y$ consists of $5421$ randomly chosen data points of every class. For the case $z\geq 30$, the size of $Y$ is appropriately adjusted in order that all the instances of class $c$ \eat{(from the MNIST training set)} exactly constitute the $z\%$ of $Y$. The instances of the other $9$ classes are randomly chosen\eat{ from the MNIST training set} so that each of them account for $(100-z)/9$ percent of $Y$. 
    \item \textbf{ImageNet} \cite{imagenet15a}: we use the popular subset corresponding to ILSVRC 2012-2017 competition. We employ $2048$ dimensional deep features~\cite{he16a}. We perform unit-norm normalization of features corresponding to each image. The source set $X$ is created by randomly sampling $50\%$ of the points. The target set $Y$ is constructed as a subset of the remaining points and depends on the skew of the target class distribution. 
    \item \textbf{Letter}\footnote{\url{https://archive.ics.uci.edu/ml/datasets/Letter+Recognition}.}: it consists of $20\,000$ data points and has $26$ classes. We randomly sample $4000$ data points as the source set and the remaining data points are used to construct target sets (with different skews) as discussed above in the case of MNIST. 
    \item \textbf{USPS}: the source set consist of $7291$ data points. The target sets are constructed from the remaining $2007$ data points, as discussed above in the case of MNIST. 
    \item \textbf{Flickr} \cite{thomee16a} is the Yahoo/Flickr Creative Commons dataset consisting of descriptive tags of various real-world outdoor/indoor images. It should be noted that unlike MNIST, Letters, or USPS, Flickr is a multi-label tag-prediction dataset, i.e., each image can have multiple tags (labels) associated with it. The dataset and the image features, extracted using MatConvNet \cite{vedaldi15a}, are available at \url{http://cbcl.mit.edu/wasserstein}. The source and target sets consists of $9836$ and $9885$ data points, respectively, from $1000$ tags (labels). 
\end{itemize}

Following \cite{proto}, we use Gaussian kernels in all our experiments. The kernel-width is chosen by cross-validation from the set $\{0.1, 0.5, 1, 5, 10\}$. Our experiments are run on a machine with $6$ core Intel CPU ($3.60$ GHz Xeon) and $64$ GB RAM. As discussed in the main paper, the quality of the representative elements selected by various methods is validated by the accuracy of the corresponding nearest prototype classifier.

\section{Additional experimental results}
\label{sec:additionalexpresults}

\subsection{Results on Letter}
The experimental set up is same as the one used in Section \ref{subsec:exp-ps-same-domain}. The good performance of {\spotg} is shown in Figure \ref{fig:letter_plots}.

\subsection{Objective value comparison}
We compare the performance of {\spots} and {\spotg} algorithms on the MNIST and ImageNet datasets. We plot the evolution of the objective value with different prototypes learned by {\spots} and {\spotg}. The plots are shown in Figure \ref{fig:objective_mnist}.

\subsection{Results on Flickr}
Since Flickr is a multi-label dataset we report an accuracy metric, where a correct prediction is assigned if and only if one of the labels from the nearest labelled image (that is used for prediction) belongs to the set of ground-truth labels corresponding to the test image. Though the metric for prediction accuracy could appear to be conservative, it is worth emphasizing that in the backdrop $1000$ possible different labels with an average of $5$ labels per data point, a random nearest neighbour assignment will lead to correct prediction only with a probability of $0.0248$ or accuracy of $\approx 2.5\%$.  Figure~\ref{fig:flickr} shows the result on the Flickr dataset. We observe that the proposed {\spotg} algorithm obtains the best result here as well.